\documentclass[10pt]{article}
\usepackage[utf8]{inputenc}
\usepackage{amssymb}
\usepackage{amsmath}
\usepackage{epsfig}
\usepackage{graphicx}

\usepackage{amsthm}
\usepackage{amsmath}
\usepackage{amssymb}
\usepackage{bm}
\usepackage{mathrsfs}

\usepackage{algorithm}
\usepackage{algorithmic}

\usepackage{color}
\usepackage[dvipsnames]{xcolor}

\usepackage{url}

\usepackage{supertabular}

\newtheoremstyle{mystyle}
  {}
  {}
  {\itshape}
  {}
  {\bfseries}
  {.}
  { }
  {}

\theoremstyle{mystyle}

\newtheorem{proposition}{Proposition}

\theoremstyle{definition}

\theoremstyle{remark}

\numberwithin{thm}{section}

\DeclareMathAlphabet{\mathsfsl}{OT1}{cmss}{m}{sl}

\renewcommand{\phi}{\varphi}

\newcommand{\argmin}{\operatorname*{arg\,min}}
\newcommand{\argmax}{\operatorname*{arg\,max}}

\newcommand{\bx}{\boldsymbol{x}}

\newcommand{\bW}{\boldsymbol{W}}

\newcommand{\by}{\boldsymbol{y}}

\newcommand{\bF}{\boldsymbol{F}}

\newcommand{\bN}{\boldsymbol{N}}

\newcommand{\bD}{\boldsymbol{D}}

\newcommand{\bU}{\boldsymbol{U}}
\newcommand{\bV}{\boldsymbol{V}}

\def\E{\mathbb{E}}
\def\bd{\boldsymbol{d}}

\def\bx{\boldsymbol{x}}

\def\beps{\boldsymbol{\epsilon}}

\def\bu{\boldsymbol{u}}

\def\bLam{\boldsymbol{\Lambda}}
\def\b0{\mathbf{0}}

\def\bU{\boldsymbol{U}}
\def\bR{\boldsymbol{R}}

\def\bv{\boldsymbol{v}}
\def\bA{\boldsymbol{A}}
\def\bY{\boldsymbol{Y}}
\def\bh{\boldsymbol{h}}
\def\br{\boldsymbol{r}}

\def\bI{\boldsymbol{I}}

\newcommand{\bSig}{\boldsymbol{\Sigma}}

\newcommand{\Cov}{\text{Cov}}
\newcommand{\Var}{\text{Var}}
\newcommand{\bmu}{\boldsymbol{\mu}}
\newcommand{\bo}{\boldsymbol{0}}

\usepackage{authblk}
\usepackage[margin=1.5in]{geometry}
\usepackage{appendix}
\usepackage{siunitx}

\author[1]{Yunpeng Shi\thanks{Corresponding author, yunpengs@princeton.edu}}
\author[1,2]{Amit Singer}
\affil[1]{Program in Applied and Computational Mathematics\\
Princeton University}
\affil[2]{Department of Mathematics\\
Princeton University}

\date{}
\begin{document}
\title{Ab-initio Contrast Estimation and Denoising of Cryo-EM Images}
\maketitle

\begin{abstract}
\noindent \textbf{Background and Objective:}\\
The contrast of cryo-EM images varies from one to another, primarily due to the uneven thickness of {\color{black} the ice layer}. {\color{black}This contrast variation} can affect the quality of 2-D class averaging, 3-D ab-initio modeling, and 3-D heterogeneity analysis. 
Contrast estimation is currently performed during 3-D iterative refinement. As a result, the estimates are not available {\color{black}at the earlier computational stages of} class averaging and ab-initio modeling. This paper aims to solve {\color{black} the contrast estimation problem directly from the picked particle images in the ab-initio stage, without estimating the 3-D volume, image rotations, or class averages.}
\\
\textbf{Methods:} \\
The key observation underlying our analysis is that the 2-D covariance matrix of the raw images is related to the covariance of the underlying clean images, the noise variance, and the contrast variability between images. We show that the contrast variability can be derived from the 2-D covariance matrix and we apply the existing Covariance Wiener Filtering (CWF) framework to estimate it. We also demonstrate a modification of CWF to estimate the contrast of individual images.     
 \\
\textbf{Results:} \\
Our method improves the contrast estimation by a large margin, compared to the previous CWF method. Its estimation accuracy is often comparable to that of an oracle that knows  the ground truth covariance of the clean images. The more accurate contrast estimation also improves the quality of image {\color{black}restoration} as demonstrated  in both synthetic and experimental datasets.
\\
\textbf{Conclusions:} \\
This paper proposes an effective method for contrast estimation directly from noisy images without using any 3-D volume information. It enables contrast correction in the earlier stage of single particle analysis, and may improve the accuracy of downstream processing. 
\end{abstract}

\section{Introduction}
In the past decade, single particle reconstruction (SPR) by cryo-electron microscopy (cryo-EM) has emerged as a critical technique for high resolution 3-D structure determination of macromolecules  \cite{bai2015cryo, cheng2018single,frank2017advances, sigworth2016principles,singer2018mathematics, singer2020computational}. In SPR, the 3-D structure needs to be determined from many noisy tomographic projection images with unknown viewing directions. Cryo-EM images are typically very noisy due to the limited electron dosage required to avoid significant radiation damage. 

Mathematically, the formation of cryo-EM images can be summarized as follows.
 Let $\phi(\boldsymbol{r})$ be the electrostatic potential of a molecule where $\br=(x, y, z)^T \in \mathbb R^3$. The $i$-th observed raw image $\bI_i$ is modeled as
    \begin{align}\label{eq:generate_3d}
    \bI_i(x, y) = c_i\bh_i \ast \int \phi(\bR_i^{-1}\br) d z + \bN_i. 
\end{align}
Namely, the molecule $\phi$ is first rotated by the rotation matrix $\bR_i$, and followed by projection in the $z$-direction to form the 2-D clean projection image. Next, the clean image is convolved with the 2-D filter $\bh_i$, often known as the point spread function, or the inverse Fourier transform of the contrast transfer function (CTF). The convolved $i$-th clean image is further rescaled by its amplitude contrast $c_i$. At last, additive noise $\bN_i$ is applied to the resulting image (translations are omitted from \eqref{eq:generate_3d} just for the sake of simplicity of exposition).
The goal of SPR is to estimate $\phi$ from the set of observed noisy images $\{\bI_i\}_{i\in [n]}$, where $[n]:=\{1,2,\dots, n\}$. 

The challenges of SPR lie in several different aspects. First, the noise term $\bN_i$ typically has much larger magnitude than that of the clean signal, making the clean signal hard to distinguish from the noise even by {\color{black}the naked eye}. As a result, a large number of particle images  ($10^4$-$10^6$) is {\color{black} often} required for reconstruction \cite{cheng2018single}. Second, the rotations $\{\bR_i\}_{i\in [n]}$ are unknown. These additional unknown variables make the estimation of the 3-D volume  difficult in the low signal-to-noise-ratio (SNR) regime. The third challenge arises from {\color{black}the} CTF. Although the CTF can be estimated from the power spectrum of the micrograph \cite{heimowitz2020reducing, penczek2014cter, rohou2015ctffind4, zhang2016gctf}, CTF correction is a challenging deconvolution problem. The main reason is that the CTFs are highly oscillatory and have zeros at many frequencies. Those zero-crossings completely remove the information of the images at those frequencies. As a result,  accurate CTF correction requires the usage of several images from different defocus groups (namely different CTFs), assuming that those CTFs have non-overlapping zero-crossings \cite{CWF}. Last but not least, the underlying clean signals may have different scaling $c_i$. This amplitude variation is mainly due to the unevenness of the ice layers where the molecule samples reside \cite{vulovic2013image}. Thicker ice layers increase inelastic scattering of electrons by ice, hence decreasing elastic scattering by the molecule and effectively weakening the signal, i.e., a smaller scaling coefficient $c_i$. The large variation of $c_i$ may cause inaccurate image denoising and CTF correction. Moreover, the scale variations may severely affect the similarity measures used to detect images from similar viewing directions for 2-D class averaging \cite{zhao2014rotationally, bhamre2017mahalanobis}, 3-D heterogeneity analysis, and the identification of common lines for 3-D ab-initio modeling \cite{bandeira2020non}. {\color{black}In particular, as pointed in Table 2 of \cite{bandeira2020non}, the uneven image contrast is the most important factor that negatively affects the accuracy of rotation estimation by some common line based approaches. At last,} scaling variability must be accounted for 3-D heterogeneity analysis to prevent artificial classes {\color{black} that correspond to contrast variations} \cite{moscovich2020cryo}. In this work, we aim to address the last challenge, contrast estimation, in the \emph{ab-initio} stage. In other words, we are interested in the direct estimation of $c_i$ without estimating $\phi$ and $\bR_i$. {\color{black} Furthermore, we use the improved contrast estimation to obtain better denoising and CTF correction of the images.}

\section{{\color{black}Related Work}}
There exist several works that estimate the amplitude contrast using  estimated $\phi$ and $\bR_i$'s \cite{scheres2012relion, sigworth2010introduction}. Specifically, assuming accurately estimated CTFs and given the {\color{black}estimates} $\widehat\phi$, $\widehat\bR_i$, one can compute the estimated $i$-th CTF-effected clean image $\widehat \bI_i := \widehat \bh_i \ast \int \widehat\phi(\widehat\bR_i^{-1}\br) d z$ and then $c_i$ can be estimated as $\langle \bI_i\,, \widehat\bI_i \rangle/ \|\widehat\bI_i\|^2$. The estimates of $c_i$, $\bR_i$ and $\phi$ are often iteratively refined using the EM algorithm \cite{scheres2009maximum}. Estimating $c_i$ without {\color{black} any} knowledge of rotations and 3-D structure is a challenging task. We refer to this task of contrast estimation as \emph{ab-initio} contrast estimation (ACE). To the best of our knowledge, ACE has not been extensively studied in previous works.  The mean pixel value of the CTF-corrected and denoised images can be used to approximate the contrast. However, \cite{CWF} only uses the estimated contrasts to filter out junk particles (outliers), while the accuracy of contrast estimation itself was not tested. There are other contrast-related techniques. Image normalization \cite{scheres2009maximum, sorzano2004normalizing} rescales the images so that the background noise level is approximately the same across the images. However, its normalization factor depends on the noise level, not the amplitude contrast. There are also works on contrast enhancement \cite{wu2018algorithm,palovcak2020enhancing,spilman2015boosting}. These aim to enhance the brightness of the underlying signal so that it is more distinguishable from the noise. However, they do not directly estimate the amplitude contrast of the clean signal, and in the process they alter the image contrast.

{\color{black}
There are also several commonly used ab-initio methods for simultaneous image denoising and CTF correction, such as traditional Wiener filtering (TWF), and covariance Wiener filtering (CWF) \cite{CWF}. TWF denoises each image using its own information, which suffers from low SNR and zero-crossings in CTF. CWF overcomes these issues by estimating the population covariance of a set of images. However, its denoising performance degrades when the covariance is not accurately estimated. Image restoration can also be done by 2-D class averaging \cite{zhao2014rotationally, multi-freq, landa2018steerable, scheres2012relion}. These methods require pairwise comparison and alignment of images, unlike the preprocessing methods such as \cite{CWF} and \cite{chung2020pre}. It is also shown in \cite{chung2020pre} that an appropriate image preprocessing can significantly improve the results of 2-D class averaging.
Deep learning based methods were recently introduced for image denoising and enhancement \cite{palovcak2020enhancing, N2N_bepler2020topaz, GAN_gu2020robust, GAN_su2018generative}. Noise2noise \cite{N2N_bepler2020topaz} requires multiple video frames of the same micrograph, which are not always available. Other CNN and GAN based methods \cite{palovcak2020enhancing, GAN_gu2020robust, GAN_su2018generative} require clean projections to form clean-noisy pairs of images to train the model, but the clean projections are not available in the ab-initio reconstruction stage, and training with clean projections of other molecules may introduce model bias.
}

\section{\color{black}Methodology}
In this work, we directly estimate the amplitude contrast from the CTF-corrected and denoised images in the ab-initio stage of SPR. Our method is based on CWF with additional constraints on the covariance matrix which we find realistic and useful for contrast estimation.

In order to address the ACE problem, we first derive from \eqref{eq:generate_3d} a simplified image formation model that is independent of $\bR_i$ and $\phi$. We then propose our method to solve the ACE problem under this model.

\subsection{A Simplified Image Formation Model}
To demonstrate the simplified model, we first reshape the images in \eqref{eq:generate_3d} as vectors and obtain
\begin{align}\label{eq:generate_2d}
        \by_i = c_i\bA_i\bx_i + \beps_i
\end{align}
where $\by_i$ and $\bx_i$ are respectively the vectors of $i$-th noisy and clean images, $\bA_i$ is the square matrix operator corresponding to the convolution with $\bh_i$, $\beps_i$ is the Gaussian noise vector and $c_i$ is the contrast to be estimated. {\color{black}
In this model, $c_i$, $\bx_i$ and $\beps_i$ are unknown. However, the power spectral density (PSD) of $\beps_i$ is assumed known as it can be estimated from the corners of the observed images $y_i$. We assume that $\bA_i$ and its Fourier transform, the CTF, are known, since they can often be accurately estimated in advance from the noisy micrographs. Throughout this work, we assume that the CTFs are radially symmetric by ignoring astigmatism.
 Without loss of generality (WLOG)} we assume that the noise distribution is white Gaussian whose covariance is $\sigma^2\bI$. For colored Gaussian noise, one can whiten the noise by applying $\bW$ (noise covariance to the power $-1/2$) to $\by_i$, so that $\bW\by_i = c_i\bW\bA_i\bx_i + \bW\beps_i$ and the {\color{black}covariance of the whitened noise $W\epsilon_i$ is the identity matrix.}  The goal of ACE is to estimate $c_i$ from the observed $\by_i$. 

Without additional assumptions on $\bx_i$ and $c_i$, the ACE problem is ill-posed due to the scale ambiguity of $\bx_i$ and $c_i$. To make it a well-posed problem, WLOG we assume that $\bx_i$ and $c_i$ are random variables such that for all $i\in [n]$,
\begin{enumerate}
    \item $c_i$ and $\bx_i$ are independent of each other. \label{assp:ind}
    \item $\E(c_i) = 1${\color{black}. } \label{assp:Ec}
    \item $\bx_i^\top \mathbf{1} = s$ for some constant $s>0$, where $\mathbf{1}$ is the {\color{black}all-ones} vector of the same {\color{black} size as} $\bx_i$. \label{assp:x1}
\end{enumerate}
The first assumption is reasonable since the contrast $c_i$ primarily depends on the thickness of the ice layer, which is indeed independent of the rotation $\bR_i$ and consequently independent of $\bx_i$. 
The second assumption is needed to overcome the global scale ambiguity of $c_i$. The last assumption states that the sum of pixel values of the clean projection image {\color{black}is the same for all clean images. This is a reasonable assumption, because} for each $i\in [n]$, the sum of the elements in $\bx_i$ is approximately $\int\int (\int \phi(\bR_i^{-1}\br) dz) dx dy = \int \phi(\br) d \br$ which is a constant independent of $i$. In other words, the sum of pixel values of any 2-D projection image equals the sum of 3-D voxel values. In fact, it is also invariant to translations (i.e., non-perfect centering of the images).  

{\color{black}We note that our model assumes that $c_i$'s are identically distributed, but it does not require the contrasts to be independent of each other. Namely, we allow correlations among $c_i$'s, which is often observed in experimental data. For example, two particle images that are closely located in the same micrograph often share similar contrasts. In principle, this information can be used to improve the contrast estimation, but this is left to  future work. We refer the reader to Figure \ref{fig:relion_coord_10028} in Section \ref{sec:10028} for further discussion.} 

We remark that the estimation of $c_i$ remains challenging due to the CTF that affects the sum of pixel values and {\color{black}due to} the high noise level. Thus, in order to obtain a good estimate of $c_i$, it is useful to denoise the image and to correct the CTF effect.
A well-known method for such image restoration is CWF \cite{CWF}, which {\color{black}is elaborated} in the next subsection.

\subsection{Preliminaries: Covariance Wiener Filtering}
CWF estimates $c_i\bx_i$ from  $\by_i$ by minimizing the expected mean squared error given an estimated covariance matrix of $c_i\bx_i$. Assume that the true covariance matrix of $c_ix_i$, denoted by $\bSig_{cx}$, is given by an oracle. Then, under the model $\by_i = \bA_i (c_i\bx_i) + \beps_i$, the linear minimum mean squared error (LMMSE) estimator is given by
\begin{align}\label{eq:cwf}
    \widehat{c_i\bx_i}^{\text{CWF}} &=  \text{CWF}(\by_i, \bA_i, \bSig_{cx}) = \argmin_{\widehat{c_i\bx_i}} \E(\|\widehat{c_i\bx_i}-c_i\bx_i\|^2 | \by_i) \\
    &=  \bmu + \bSig_{cx} \bA^\top_i(\bA_i\bSig_{cx} \bA_i^\top + \sigma^2 \boldsymbol{I})^{-1} (\by_i - \bA_i\boldsymbol\mu),
\end{align}
where $\bmu$ is the true mean of $c_i\bx_i$.

We note that CWF naturally induces an optimal linear estimator of contrast given the true covariance $\bSig_{cx}$. Indeed, it can be easily shown  that {\color{black}:}
\begin{align}\label{eq:cwf_contrast}
     &\argmin_{\mathbf{1}^\top\widehat{c_i\bx_i}} \E(\|\mathbf{1}^\top\widehat{c_i\bx_i}-\mathbf{1}^\top c_i\bx_i\|^2 | \by_i) \nonumber\\
    =&  \mathbf{1}^\top\bmu + \mathbf{1}^\top\bSig_{cx} \bA^\top_i(\bA_i\bSig_{cx} \bA_i^\top + \sigma^2 \boldsymbol{I})^{-1} (\by_i - \bA_i\boldsymbol\mu) = \mathbf{1}^\top \widehat{c_i\bx_i}^{\text{CWF}}. 
\end{align}
Namely, $\mathbf{1}^\top \widehat{c_i\bx_i}^{\text{CWF}}$, the sum of pixel values of the CWF estimate of $c_i\bx_i$, is the best linear estimator of $\mathbf{1}^\top c_i\bx_i$ given $\bSig_{cx}$. Note that  $\mathbf{1}^\top c_i\bx_i=s\cdot c_i $ by the third assumption of our model. Therefore, we have obtained the optimal linear estimator of contrast $c_i$ up to a global constant $s$. This scale ambiguity can be solved by using the second assumption $\E(c_i)=1$ in our model. That is, after obtaining the estimates of $s\cdot c_i$, we normalize the estimates by a global constant so that the average of the set $\{\mathbf{1}^\top \widehat{c_i\bx_i}^{\text{CWF}}\}_{i\in [n]}$ is 1.

However, the optimal properties of the aforementioned estimates only hold when $\bSig_{cx}$ is given, which is not true in practice. In \cite{CWF}, the mean of $c_i\bx_i$ is estimated by minimizing the least squares error between the noisy images and the CTF transformed mean. Specifically, 
\begin{align}\label{eq:est_mean}
   \widehat\bmu = \argmin_{\bmu} \sum_{i=1}^n\|\by_i - \bA_i\bmu\|^2.
\end{align}
Similarly, the covariance matrix $\bSig_{cx}$ is estimated by  minimizing the {\color{black}squared deviations}  between the sample covariance of $\by_i$ and the population covariance of $\bA_i(c_i\bx_i) + \beps_i$. 
That is,
\begin{align}\label{eq:cov}
   \widehat\bSig_{cx} =  \argmin_{\bSig} \sum_{i=1}^n \left\|(\by_i - \bA_i\widehat\bmu)(\by_i - \bA_i\widehat\bmu)^\top - (\bA_i\bSig\bA_i^\top+\sigma^2 \boldsymbol{I})\right\|_F^2.
\end{align}
By setting the first order derivative of \eqref{eq:cov} to zero, we end up with the following linear system of equations:
\begin{align}\label{eq:cov_linear}
    \sum_{i=1}^n\bA_i^\top \bA_i {\color{black}\widehat{\bSig}_{cx}} \bA_i^\top \bA_i = \sum_{i=1}^n\bA_i^\top (\by_i - \bA_i\widehat\bmu)(\by_i - \bA_i\widehat\bmu)^\top \bA_i + \sigma^2 \sum_{i=1}^n\bA_i^\top \bA_i.
\end{align}
We note that the first term {\color{black} on the right hand side (RHS)} of \eqref{eq:cov_linear} corresponds to the sample covariance of $\bA_i^\top\by_i$. However, $\by_i$ often has dimension $>10^4$ which is comparable to the number of images. In this high dimensional setting, the sample covariance is not a consistent estimator of the population {\color{black}covariance}. As a result, an eigenvalue shrinkage method is applied to the RHS of \eqref{eq:cov_linear} to improve the covariance estimation. At last, \eqref{eq:cov_linear} is solved by applying the conjugate gradient method. We refer the readers to \cite{CWF} for more details.

We remark that under low SNR or insufficient number of samples, $\bSig_{cx}$ could be poorly estimated. In such a case, the CWF method and its induced contrast estimator \eqref{eq:cwf_contrast} are far from being optimal. Therefore, there is still room for improvement on the CWF-based contrast estimation. Indeed, the CWF-estimator does not fully exploit our model assumptions. As we show in the next subsection, the three assumptions of our model imply novel constraints on $\bSig_{cx}$ which turn out to significantly improve contrast estimation.

\subsection{Novel Constraints on the Covariance Matrices}
The new constraints on the covariance matrix are stated in the following proposition.
Let $\bSig_x$ be the true covariance of $\bx_i$ and  ${\color{black}\Var(c)}$ be the variance of each $c_i$.
\begin{proposition}\label{prop:constr}
If the three assumptions for the model \eqref{eq:generate_2d} are satisfied, then the following two constraints hold:
\begin{align}
     &1. \quad \bSig_{cx} = ({\color{black}\Var(c)}+1)\bSig_x + {\color{black}\Var(c)}\bmu\bmu^\top \label{eq:assump1}\\
    &2.\quad \bSig_{x}\mathbf{1} = \boldsymbol{0} \label{eq:assump2}.
\end{align}
\end{proposition}
This proposition suggests that the true covariance of $c_i\bx_i$ is the combination of two components, one corresponds to the covariance without contrast variability whose eigenvectors are perpendicular to the {\color{black}all-ones} vector, and the other corresponds to {\color{black}a} rank-one matrix whose eigenvector is the mean.
The derivation of the two constraints is simple. To prove the first constraint,
we use the identity that for any independent scalar random variable $c$ and random vector $\bx$, $\Cov(c\bx) = \E(c^2)\Cov(\bx) + {\color{black}\Var(c)}\E(\bx)\E(\bx)^\top$. 
By letting $c=c_i$ and $\bx=\bx_i$, we obtain that for any $i\in [n]$
\begin{align*}
   \bSig_{cx} &= {\color{black}\Cov(c\bx) = \E(c^2)\Cov(\bx) + {\color{black}\Var(c)}\E(\bx)\E(\bx)^\top}\\
   & = ({\color{black}\Var(c)}+\E^2(c))\bSig_x + {\color{black}\Var(c)}\bmu\bmu^\top.
\end{align*}
By using the assumption $\E(c_i)=1$, we conclude the first constraint.
The second constraint states that the variation of the sum of elements in $\bx_i$ is 0, namely the contrast variability of clean signals is 0. It can be verified easily by using the third assumption of our model. Specifically,
\begin{align*}
    \bSig_x \mathbf{1} = \E\left((\bx_i-\bmu)(\bx_i-\bmu)^\top\right) \mathbf{1} = \E\left((\bx_i-\bmu)(\bx_i^\top \mathbf{1}-\bmu^\top \mathbf{1})\right)  = \bo,
\end{align*}
where the last equality follows from the assumption that {\color{black}:} $\bx_i^\top \mathbf{1}=\bmu^\top \mathbf{1}=s$. In the next subsection, we propose two methods that use the two covariance constraints to refine the estimated covariance $\bSig_{cx}$.

\subsection{Refinement of Covariance Matrices}
We first use the two covariance constraints {\color{black}\eqref{eq:assump1} and \eqref{eq:assump2}} to estimate the contrast variance ${\color{black}\Var(c)}$. By combining the two constraints,
\begin{align}\label{eq:sigma_cx_1}
    \bSig_{cx}\mathbf 1 = ({\color{black}\Var(c)}+1)\bSig_x \mathbf 1 + {\color{black}\Var(c)}\bmu\bmu^\top \mathbf 1 = {\color{black}\Var(c)}\bmu\bmu^\top \mathbf{1},
\end{align}
where the second equality follows from the second covariance constraint.
We note that \eqref{eq:sigma_cx_1} relates ${\color{black}\Var(c)}$ to $\bSig_{cx}$ and $\bmu$, where the latter two can be estimated from the noisy data. Given the estimated $\widehat{\bSig}_{cx}$ and $\widehat{\bmu}$, the variance of the image contrast can be estimated by least squares as follows.
\begin{align}\label{eq:var}
    \widehat{\color{black}\Var(c)} = \argmin_{{\color{black}\Var(c)}}\|\widehat\bSig_{cx}\mathbf 1 -  {\color{black}\Var(c)} (\widehat\bmu^\top \mathbf{1}) \widehat\bmu\|_2^2 
    = \frac{\widehat\bmu^\top \widehat\bSig_{cx} \mathbf 1}{\|\widehat\bmu\|^2\widehat\bmu^\top \mathbf{1}}.
\end{align}

We remark that the {\color{black} initially} estimated $\widehat\bSig_{cx}$ often does not satisfy the constraints in Proposition \ref{prop:constr}. Therefore, we introduce two methods to enforce the covariance constraints {\color{black} using the estimated $\widehat{\color{black}\Var(c)}$}. We refer to the first method as semi-definite programming (SDP).
\\
\\
\textbf{SDP:} We seek to find the closest positive semidefinite matrix to the initially estimated $\widehat\bSig_{cx}$ such that the two covariance constraints are satisfied. Namely, we seek a solution of the following SDP problem.
\begin{align}\label{eq:sdp}
    \widehat\bSig_{cx}^{\text{SDP}}& = \argmin_{\bSig_{cx}^{\text{SDP}}} \left\|\bSig_{cx}^{\text{SDP}}-\widehat\bSig_{cx} \right\|_F^2 \\\nonumber
    \text{subject to} \quad 
    \bSig_{cx}^{\text{SDP}} &= ({\color{black} \widehat{\Var(c)}}+1)\bSig_x^{\text{SDP}} + {\color{black} \widehat{\Var(c)}}\widehat\bmu\widehat\bmu^\top\\\nonumber
    \bSig_{cx}^{\text{SDP}} &\succeq \boldsymbol{0}\\\nonumber
    \bSig_x^{\text{SDP}} &\succeq \boldsymbol{0}\\\nonumber
    \bSig_x^{\text{SDP}}\mathbf{1} &=\boldsymbol{0}.
\end{align}
Since $\bSig_{cx}^{\text{SDP}} = ({\color{black} \widehat{\Var(c)}}+1)\bSig_x^{\text{SDP}} + {\color{black} \widehat{\Var(c)}}\widehat\bmu\widehat\bmu^\top$, $\bSig_{cx}^{\text{SDP}}$ is positive semidefinite when $\bSig_x^{\text{SDP}}$ is so. Thus, one can drop the constraint $\bSig_{cx}^{\text{SDP}} \succeq \boldsymbol{0}$, and plug in the first constraint of \eqref{eq:sdp} to $\bSig_{cx}^{\text{SDP}}$ in its objective function. This yields the following SDP fomulation that optimizes for $\bSig_{x}^{\text{SDP}}$.
\begin{align}\label{eq:sdp_x}
    \widehat\bSig_x^{\text{SDP}} = \argmin_{\bSig_x^{\text{SDP}}} &\left\|\bSig_x^{\text{SDP}}- \frac{\widehat\bSig_{cx} - \widehat{\color{black}\Var(c)}\widehat\bmu\widehat\bmu^\top}{\widehat{\color{black}\Var(c)}+1}\right\|_F^2 \\\nonumber
    \text{subject to}& \quad  \bSig_x^{\text{SDP}}\mathbf{1}=\bo\\\nonumber
    & \quad \bSig_{x}^{\text{SDP}} \succeq \boldsymbol{0}.
\end{align}
After solving $\widehat\bSig_{x}^{\text{SDP}}$, we immediately obtain $\widehat\bSig_{cx}^{\text{SDP}} = ({\color{black} \widehat{\Var(c)}}+1)\widehat\bSig_x^{\text{SDP}} + {\color{black} \widehat{\Var(c)}}\widehat\bmu\widehat\bmu^\top$. Next, we introduce a faster but heuristic alternative that uses the Gram-Schmidt (GS) process to approximately solve \eqref{eq:sdp_x}.
\\
\\
\textbf{Gram-Schmidt (GS) Process:}
We first note that the constraint $\bSig_x^{\text{SDP}}\mathbf{1}=\boldsymbol{0}$ in \eqref{eq:sdp_x} is equivalent to that all the eigenvectors of $\bSig_x^{\text{SDP}}$ are orthogonal to $\mathbf{1}$.
Similar to \eqref{eq:sdp_x}, we seek a positive semidefinite matrix $\bSig_x^{\text{GS}}$ that is close to $\widehat\bSig_x := (\widehat\bSig_{cx} - \widehat{\color{black}\Var(c)}\widehat\bmu\widehat\bmu^\top)/(\widehat{\color{black}\Var(c)}+1)$ whose eigenvectors are orthogonal to $\mathbf{1}$.

Let $\widehat\bSig_x = \widehat\bV \widehat\bD \widehat\bV^\top$ be the eigenvalue decomposition of $\widehat\bSig_x$, where $\widehat\bV$ is the eigenmatrix whose columns are eigenvectors of $\widehat\bSig_x$, and $\widehat\bD$ is the diagonal matrix of its eigenvalues. 
We seek a refined covariance $\widehat\bSig_x^{\text{GS}} = \widehat\bU \widehat\bLam \widehat\bU^\top$ with refined eigenvalues and eigenvectors such that $\widehat\bLam$ is nonnegative (so that $\widehat\bSig_x^{\text{GS}}\succeq \boldsymbol{0}$) and  $\widehat \bU^\top\mathbf{1}=\boldsymbol{0}$ (so that $(\widehat\bSig_x^{\text{GS}})^\top\mathbf{1}=\boldsymbol{0}$), and $\widehat \bLam$ and $\widehat \bU$ are respectively close to $\widehat\bD$ and $\widehat\bV$.

The solution of $\widehat \bLam$ is obtained by simply thresholding the negative values in $\widehat\bD$. That is, $\widehat \bLam = \max(\widehat\bD, 0)$. The solution of the eigenmatrix $\widehat\bU$ is trickier, due to the nonconvex constraint $\widehat\bU^\top \widehat\bU = \bI$, namely the columns of $\widehat\bU$ (the eigenvectors of $\widehat\bSig_x^{\text{GS}}$) form an orthonormal basis. It asks to solve the following nonconvex optimization problem.
\begin{align}\label{eq:optV}
    \widehat\bU = \argmax_{\bU} &\,\,\text{Tr}(\widehat\bV^\top \bU) \\\nonumber
    \text{subject to}\quad  \bU^\top \bU &= \bI \\\nonumber
     \bU^\top\mathbf{1}&=\boldsymbol{0}.
\end{align}
Instead of directly solving \eqref{eq:optV}, we argue that a simple Gram-Schmidt process on $\widehat\bV$ is often sufficient to obtain a satisfying solution. Let $\widehat\bV = [\widehat\bv_1, \widehat\bv_2, \dots, \widehat\bv_p]$ where $p$ is the dimension of each $\bx_i$, and the eigenvectors are placed in  descending order of eigenvalues. 

Let $\widehat\bU := [\widehat\bu_1, \widehat\bu_2, \dots, \widehat\bu_p]$ and $[\mathbf{1}, \widehat\bV_{-p}]:=[\mathbf{1}, \widehat\bv_1, \widehat\bv_2, \dots, \widehat\bv_{p-1}]$ be $p$-by-$p$  square matrices. Application of Gram-Schmidt orthogonalization to $[\mathbf{1}, \widehat\bV_{-p}]$ yields a new orthogonal matrix $[\mathbf{1}, \widehat{\bU}_{-p}]:=[\mathbf{1}, \widehat\bu_1, \widehat\bu_2, \dots, \widehat\bu_{p-1}]$. That is, $\widehat\bu_1$ is computed by projecting $\widehat\bv_1$ onto the orthogonal complement of $\mathbf{1}$ and then normalize to unit vector. Once $\widehat \bu_{i-1}$ for $1\leq i\leq p-1$ are computed, $\widehat\bu_i$ is computed by projecting $\widehat\bv_i$ onto the orthogonal complement of linear subspace spanned by $\mathbf{1}, \widehat\bu_1, \dots, \widehat\bu_{i-1}$ and then normalize.  At last, the solution $\widehat\bU$ is obtained by finding the orthogonal complement of $\widehat\bU_{-p}$ to complete its missing column $\widehat\bu_{p}$. In this way, the columns of  $\widehat\bU$ form an orthonormal basis, and its first $p-1$ columns are orthogonal to $\mathbf{1}$. Although $\widehat\bu_p$ may not necessarily be orthogonal to $\mathbf{1}$, its eigenvalue is 0 in most of the cases and thus won't affect the solution of $\widehat\bSig_x^{\text{GS}}$. {\color{black}In practice, the GS process is done by the QR decomposition for its better numerical stability.}

 Iterating from the top eigenvectors has two benefits. First, the top eigenvectors of $\widehat\bSig_x$ are more robust to the noise. That is, the top eigenvectors $\widehat\bv_1, \widehat\bv_2, \dots$ of $\widehat\bSig_x$ are often closer to those of the true $\bSig_x$. Due to the constraint $\bSig_x\mathbf{1}=\boldsymbol{0}$, the top eigenvectors of $\widehat\bSig_x$ often have smaller correlation with $\mathbf{1}$. In other words, the top eigenvectors are cleaner and thus their refinement is easier and more accurate, and therefore they should be put at the earlier stage of the sequential projection procedure to reduce error accumulation. Second, iterating from the top eigenvectors makes them more accurately projected {\color{black} to the orthogonal complement of $\mathbf{1}$ with minimal changes to their original values}.  This is beneficial for contrast estimation since these top eigenvectors are more important for explaining the contrast variations.

\subsection{Ab-initio Contrast Estimation and Denoising}\label{sec:denoise}

After applying the aforementioned SDP or GS method to the initial covariance matrix $\widehat\bSig_{cx}$, we obtain the refined covariance $\widehat\bSig_{cx}^{\text{RF}}=\widehat\bSig_{cx}^{\text{SDP}}$ or $\widehat\bSig_{cx}^{\text{GS}}$. 

Recall that the CWF estimator of $c_i\bx_i$ is defined in \eqref{eq:cwf}. Our refined estimate of $c_i\bx_i$ for each $i\in [n]$ is
\begin{align}\label{eq:cwf_ci}
    \widehat{c_i\bx_i}^{\text{RF}}=\text{CWF}\left(\by_i, \bA_i, \widehat\bSig_{cx}^\text{RF}\right).
\end{align}
Then, by applying our model assumptions that $\mathbf{1}^\top\bx_i=s$ for all $i\in [n]$ and $\E(c_i)=1$, we obtain our refined estimate of $c_i$ as
\begin{align}\label{eq:c_rf}
    \widehat{c_i}^{\text{RF}} = \frac{\mathbf{1}^\top\widehat{c_i\bx_i}^{\text{RF}}}{\frac{1}{n}\sum_{i=1}^n \mathbf{1}^\top\widehat{c_i\bx_i}^{\text{RF}}}{\color{black},}
\end{align}
where the numerator is the contrast estimator up to $s$. The denominator is the normalization factor to remove the scale ambiguity and enforce the average of $\widehat{c_i}^{\text{RF}}$ to be $1$.

After estimating the contrasts, we present two methods for estimating the clean image $\bx_i${\color{black}:} the image normalization and 2-stage CWF.\\
\\
\textbf{Image Normalization}\\
In the first approach, we estimate $\bx_i$ as 
\begin{align}\label{eq:img_norm}
    \widehat{\bx}_i^{\text{RF}} = \frac{\widehat{c_i\bx_i}^{\text{RF}}}{\widehat{c_i}^{\text{RF}}}.
\end{align}
That is, we simply normalize the estimated $c_i\bx_i$ by the estimated contrast $c_i$, so that the resulting images all share the same sum of pixel values.
\\
\\
\textbf{2-Stage CWF}\\
In the second method, we apply an additional CWF estimator to directly estimate $\bx_i$. Recall that the original version of CWF aims to estimate $c_i\bx_i$. It treats $c_i\bx_i$ as a single variable and considers the model $\by_i = \bA_i(c_i\bx_i) + \beps_i$.  In order to directly estimate $\bx_i$, we treat $c_i$ as known and absorb it into the CTF term, and consider the model $\by_i = (\widehat{c_i}^{\text{RF}}\bA_i) \bx_i + \beps_i$. Given this model, a natural estimate of $\bx_i$ is
\begin{align}\label{eq:2stage}
    \widehat{\bx}_i^{\text{RF}} = \text{CWF}\left(\by_i,\, \widehat{c_i}^{\text{RF}}\bA_i,\, \widehat\bSig_{x}^\text{RF}\right).
\end{align}
Since the refined $\widehat\bSig_{x}^\text{RF}$ satisfies the constraint $\widehat\bSig_{x}^\text{RF}\mathbf{1} = \boldsymbol{0}$, the resulting recovered image $\widehat{\bx}_i^{\text{RF}}$ automatically {\color{black}has} the same sum of pixel values. Ideally, if $\widehat{c_i}^{\text{RF}} = c_i$ and $\widehat\bSig_{x}^\text{RF} = \bSig_{x}$, then \eqref{eq:2stage} is the optimal linear estimator of $\bx_i$.

\subsection{Computational Issues and Steerable Basis}
Although our model and methodology {\color{black}were presented} in real image space for simplicity, in practice, implementing the CWF-based methods in real image domain is computationally intractable and memory demanding. For images of size $L \times L$, the  dimension of $\bx_i$ is of order $O(L^2)$, and the covariance of $\bx_i$ in real space has $O(L^4)$ entries. This leads to high time and space complexities that make the computation impractical.

Therefore, we follow \cite{CWF} and expand the Fourier transformed images $\mathcal F(\bI_i)$ in the Fourier-Bessel basis
\begin{align}
    \psi^{k,q}_r(\theta, \xi) =
    \begin{cases} N_{k,q} J_k(R_{k,q}\xi/r) e^{\i k\theta} & \xi\leq r \\
    0 & \text{otherwise},
    \end{cases}
\end{align}
where $0<r\leq 1/2$ is the band-limit radius of images (default $=1/2$), $k,q$ are respectively angular and radial frequencies, $(\xi, \theta)$ are the polar coordinates in Fourier domain, $J_k$ is the Bessel function of the first kind of order $k$, $R_{k,q}$ is the $q$-th root of $J_k$, and $N_{k,q}=(r\sqrt{\pi}|J_{k+1}(R_{k,q})|)^{-1}$ is the normalization factor. We refer the readers to \cite{zhao2016fast} for details of the expansion.

Denote the image formation model in the Fourier-Bessel basis as
\begin{align*}
    \by_i^{\text{FB}} = c_i\bA_i^{\text{FB}} \bx_i^{\text{FB}} + \beps_i 
\end{align*}
where $\bA_i^{\text{FB}}$, $\bx_i^{\text{FB}}$  $\by_i^{\text{FB}}$  are respectively the CTF, {\color{black} and the }clean and noisy Fourier {\color{black}transformed} images in the Fourier-Bessel basis. 
Expanding images in Fourier-Bessel basis (or other steerable basis) enjoys some nice properties. For example, image rotation in {\color{black}the} Fourier-Bessel domain is easy. Indeed, rotation of images corresponds to phase modulation of their corresponding Fourier-Bessel coefficients. This relationship between rotation and phase modulation enables easy and fast computation of the covariance matrix of any set of images that are augmented by all their possible in-plane rotations and reflections. More importantly, as shown in \cite{zhao2016fast}, the resulting covariance matrix of the augmented images is block diagonal, where blocks are indexed by the  angular frequency $k$. That is, the $((k_1, q_1), (k_2, q_2))$-th entry of the covariance matrix is nonzero only when the angular frequencies are equal, namely $k_1=k_2$. This reduces the number of variables in the covariance matrix from $O(L^4)$ to $O(L^3)$ which is a significant saving of computation time and memory usage. Similarly, a radially symmetric CTF in the Fourier-Bessel basis is also block diagonal and has the same block structure as that of the covariance. As a result, the estimation of each diagonal block of $\bx_i^{\text{FB}}$ is completely independent and decoupled from the rest of the blocks. Thus, the task of estimating $\bx_i^{\text{FB}}$ is divided into $O(L)$ independent tasks of much smaller sizes, which enables faster and parallelized computation.

Although the Fourier-Bessel expansion {\color{black}facilitates fast computation of CWF}, our model and covariance refinement method require more careful adaptation to the Fourier Bessel domain. The main issue is that {\color{black} the} Fourier-Bessel transform preserves the $\ell_2$ norm by Parseval's identity, but not the sum of pixel values. As a result,  $\mathbf{1}^\top \bx_i^{\text{FB}} = s$ and $\bSig^{\text{FB}}_x\mathbf{1} = \boldsymbol{0}$ are not necessarily satisfied.

To address this issue, we observe that
\begin{align}\label{eq:fb_contrast}
   \mathbf{1}^\top \bx_i = \mathbf{1}^\top \bF^*\bF_{B}^*\bF_B \bF \bx_i = \mathbf{1}_{\text{FB}}^\top \bx_i^{\text{FB}}, 
\end{align}
where $\bF, \bF_B$ are the matrix operators of {\color{black}the} Fourier and Fourier-Bessel {\color{black}transforms}, $\bF^*, \bF_B^*$ denote their corresponding adjoint {\color{black}operators}, $\mathbf{1}_{\text{FB}} = \bF_B\bF\mathbf{1}$ is the Fourier-Bessel transform of the Fourier-transformed {\color{black}all-ones} image. Therefore, the new constraints in the Fourier-Bessel domain are $\mathbf{1}_{\text{FB}}^\top \bx_i^{\text{FB}} = s$ and $\bSig^{\text{FB}}_x\mathbf{1}_{\text{FB}} = \boldsymbol{0}$. By replacing every $\mathbf{1}$ in the previous formulation  by $\mathbf{1}_{\text{FB}}$, exactly the same argument automatically follows in the Fourier-Bessel domain.

We finally remark that $\mathbf{1}_{\text{FB}}$ is only nonzero in the zero-th angular frequency. Indeed, $\bF \mathbf{1}$ is the dirac delta image $\bI_{\delta}$ whose only nonzero pixel is located at the origin. Let $\mathbf{1}_{k,q}$ be the $(k,q)$-th coefficient of $\mathbf{1}_{\text{FB}}$. 
\begin{align}
\mathbf{1}_{k,q} &= \int\int \bI_{\delta}(\theta, \xi)   \overline{\psi^{k,q}_r(\theta, \xi)}\xi d\xi d\theta = \overline{\psi^{k,q}_r(0, 0)} = N_{k,q} J_k(0) 
\\
&= \begin{cases}
 N_{0,q} = \frac{1}{r\sqrt{\pi}|J_{1}(R_{0,q})|}  & \text{for } k= 0,\\
 0 & \text{otherwise},
 \end{cases}\label{eq:1kq}
\end{align}
where the last equality follows from that{\color{black}:} $J_k(0)=1$ when $k=0$ and $J_k(0)=0$ for $k\neq 0$.
{\color{black}In view of \eqref{eq:fb_contrast} and \eqref{eq:1kq}}, the contrasts of the real images are only determined by the zero-th angular blocks of their Fourier-Bessel expansion. This is a favorable property from the computational aspect. For example, when computing the numerator of \eqref{eq:c_rf}, one can simply take the dot product between the zero-th angular frequency blocks of the Fourier-Bessel coefficients of $\mathbf{1}$ and $\widehat{c_i\bx_i}^{\text{RF}}$, without the need to access the entire vectors.

{\color{black}
\subsection{Summary of the Algorithm and Computational Complexity}\label{sec:summary}
Our ACE and image {\color{black} restoration} methods are respectively summarized in Algorithms \ref{alg:ACE} and \ref{alg:AD}. 
\begin{algorithm}[h]
\caption{{\color{black}Ab-initio Contrast Estimation}}\label{alg:ACE}
\begin{algorithmic}
{\color{black}
\REQUIRE $\{A_{i}\}_{i\in [n]}$, $\{\by_i\}_{i\in [n]}$, option=SDP or GS
\STATE $\bA_{i,0} \leftarrow$ extract the zero-th angular frequency diagonal block of $\bA_i$  for $i\in [n]$
\STATE $\by_{i,0} \leftarrow$ extract the entries of $\by_i$  corresponding to the zero-th angular frequency for $i\in [n]$
\STATE $\widehat\bmu_0 \leftarrow \argmin_{\bmu} \sum_{i=1}^n\|\by_{i,0} - \bA_{i,0}\bmu\|^2$ 
\STATE $\widehat\bSig_{cx,0} \leftarrow \argmin_{\bSig} \sum_{i=1}^n \left\|(\by_{i,0} - \bA_{i,0}\widehat\bmu_0)(\by_{i,0} - \bA_{i,0}\widehat\bmu_0)^\top - (\bA_{i,0}\bSig\bA_{i,0}^\top+\sigma^2 \boldsymbol{I})\right\|_F^2$
\STATE  $\widehat{\color{black}\Var(c)} \leftarrow \widehat\bmu^{\top}_0 \widehat\bSig_{cx,0} \mathbf 1_0/(\|\widehat\bmu_0\|^2\widehat\bmu_0^\top \mathbf{1}_0)$
\STATE $\widehat\bSig_{x,0} \leftarrow (\widehat\bSig_{cx,0} - \widehat{\color{black}\Var(c)}\widehat\bmu_0\widehat\bmu_0^\top)/(\widehat{\color{black}\Var(c)}+1)$
\IF{option=SDP}
\STATE $\widehat\bSig_{x,0}^{\text{RF}} \leftarrow \argmin_{\bSig_{x,0}^{\text{SDP}}} \left\|\bSig_{x,0}^{\text{SDP}}-\widehat\bSig_{x,0} \right\|_F^2 \quad
    \text{subject to} \quad  \bSig_{x,0}^{\text{SDP}}\mathbf{1}_0=\bo
     \quad \bSig_{x,0}^{\text{SDP}} \succeq \boldsymbol{0}$
\ELSE
\STATE $(\widehat\bd_i\,, \widehat\bv_i)_{i=1}^p\leftarrow$
 pairs of eigenvalues/eigenvectors of $\widehat\bSig_{x,0}$ sorted in descending order
\STATE $\widehat\bD_0 \leftarrow \text{diag}(\max(\widehat\bd_i, 0))$ 
\STATE $\widehat\bV_0 \leftarrow [\mathbf{1}_0, \widehat\bv_1, \widehat\bv_2, \dots, \widehat\bv_{p-1}]$
\STATE $[\mathbf{1}_0, \widehat\bU_0] \leftarrow \text{Gram-Schmidt}(\widehat\bV_0)$
\STATE $\widehat\bU_0 \leftarrow [\widehat\bU_0, \bv_p]$
\STATE $\widehat\bSig_{x,0}^{\text{RF}} \leftarrow \widehat\bU_0 \widehat\bD_0 \widehat\bU_0^\top$
\ENDIF
\STATE $     \widehat\bSig_{cx,0}^\text{RF}\leftarrow  ({\color{black} \widehat{\Var(c)}}+1)\bSig_{x,0}^{\text{RF}} + {\color{black} \widehat{\Var(c)}}\widehat\bmu_0\widehat\bmu_0^\top$
\STATE $    \widehat{c_i\bx_{i,0}}^{\text{RF}}\leftarrow\text{CWF}\left(\by_{i,0}, \bA_{i,0}, \widehat\bSig_{cx,0}^\text{RF}\right)$
\STATE $\widehat{c_i}^{\text{RF}} \leftarrow n\mathbf{1}_0^\top\widehat{c_i\bx_{i,0}}^{\text{RF}}/\sum_{i=1}^n \mathbf{1}_0^\top\widehat{c_i\bx_{i,0}}^{\text{RF}}$
\ENSURE $\{\widehat{c_i}^{\text{RF}}\}_{i\in [n]}$
}
\end{algorithmic}
\end{algorithm}

\begin{algorithm}[h]
\caption{{\color{black}Ab-initio Image Restoration}}\label{alg:AD}
\begin{algorithmic}
{\color{black}
\REQUIRE $\{A_{i}\}_{i\in [n]}$, $\{\by_i\}_{i\in [n]}$, option=normalization or 2-stage
\STATE $\widehat \mu$, $\widehat\bSig_{cx}\leftarrow$ by solving \eqref{eq:est_mean}, \eqref{eq:cov}
\STATE $\{\widehat{c_i}^{\text{RF}}\}_{i\in [n]}, \widehat\bSig_{cx,0}^{\text{RF}}, \widehat\bSig_{x,0}^{\text{RF}}\leftarrow$ by implementing Algorithm \ref{alg:ACE}
\STATE  $\widehat\bSig_{cx}^{\text{RF}} \leftarrow$ replace the zero-th angular frequency diagonal block of $\widehat\bSig_{cx}$ by $\widehat\bSig_{cx,0}^{\text{RF}}$
\STATE $\widehat\bSig_{x}^{\text{RF}} \leftarrow$ replace the zero-th angular frequency diagonal block of $\widehat\bSig_{cx}$ by $\widehat\bSig_{x,0}^{\text{RF}}$
\IF{option = normalization}
\STATE $\widehat{c_i\bx_{i}}^{\text{RF}}\leftarrow$ by \eqref{eq:cwf_ci}
\STATE $\widehat{\bx}_i^{\text{RF}}\leftarrow$ by \eqref{eq:img_norm}
\ELSE
\STATE $\widehat{\bx_{i}}^{\text{RF}}\leftarrow$ by \eqref{eq:2stage}
\ENDIF
\ENSURE $\{\widehat{\bx}_i^{\text{RF}}\}_{i\in [n]}$
}
\end{algorithmic}
\end{algorithm}
}

We comment on the computational complexity of our methods. From \cite{CWF}, the overall complexity for the original CWF is $O(TDL^4 +nL^3)$. The first term {\color{black}corresponds to covariance estimation}, where $T$ is the
number of conjugate gradient iterations for estimating $\bSig_{cx}$ and $D$ is the number of defocus groups. The second term {\color{black}corresponds to denoising by Wiener filtering.} Our contrast estimator takes two additional steps that cost extra computation. The covariance refinement by GS process takes $O(L^3)$ operations due to the eigenvalue decomposition of the diagonal block of $\bSig_x$ corresponding to the zero-th angular frequency. {\color{black}This step is negligible compared to the computational complexity of CWF.} In contrast, the SDP method suffers from much higher complexity. For both splitting conic solver (SCS) \cite{scs} and interior point method \cite{interior}, the per-iteration complexity is $O(m^3)$ where $m=L^2$ is the number of variables in the diagonal block of $\bSig_{x}$ corresponding to the zero-th angular frequency. Therefore, the total complexity of our SDP method (using the aforementioned solvers) is $O(L^6)$. Nevertheless, in our synthetic and experimental data $L<400$ so empirically we observe that SDP can still be implemented in less than one hour. {\color{black}The step of estimating the image contrasts using the Fourier-Bessel basis (eq. \eqref{eq:fb_contrast}) requires $O(nL)$ operations, which is negligible compared to the cost of CWF.} In summary, our method with GS-refinement of covariance has similar complexity to that of the original CWF. Our method with SDP-refinement of covariance has higher complexity but it is still practical.

\section{Results for Synthetic Data}
In this section, we compare our method with the original CWF method for contrast estimation and image denoising using synthetic data. To generate the synthetic data, we create the 2-D clean images by projecting a 3-D volume from uniformly distributed viewing directions. The images are downsampled to size $256\times 256$. We use the 3-D volume of the P. falciparum 80S ribosome bound to
E-tRNA, which can be freely obtained from the Electron Microscopy Data Bank (EMDB) with ID number EMD-2660 \cite{10028}. We apply 10 different CTFs to the projected clean images, whose defocus values
range from 1 {\color{black}\si{\mu m}} to 4 {\color{black}\si{\mu m}}.
For all CTFs, we choose the voltage as 300 {\color{black}\si{kV}}, the amplitude contrast as $7\%$, and the spherical
aberration as 2 {\color{black}\si{mm}}. We then rescale the clean CTF-transformed images by image amplitude contrasts that are i.i.d.~uniformly distributed in $[0.5, 1.5]$. At last, we add additive white or colored Gaussian noise. For the colored noise, we choose the noise power spectrum as $1/\sqrt{k^2+1}$ up to a constant, where $k$ is the radial frequency  {\color{black} (in 1/(128 pixel size))  in the Fourier domain. The pixel size is set as $1.34 \times 360/256$ \AA, where $360$ is the original dimension of the volume before downsampling.}

We implement all algorithms on a cluster with 750GB {\color{black} shared memory} and 72 cores running at 2.3 GHz, {\color{black}where 20 cores were used.} We implement CWF using the ASPIRE package \cite{aspire} with its default setting. As for our methods, the SDP covariance refinement formulation is solved in CVXPY \cite{cvxpy} by its default solver SCS \cite{scs}. {\color{black}Our Python code is available at} \url{https://github.com/yunpeng-shi/contrast-cryo}, {\color{black}and} is planned to be integrated into ASPIRE.

We next comment on the runtime of the algorithms. The Fourier-Bessel expansion for a batch of 1000 images {\color{black}takes} 110 seconds. With 10 defocus groups, the covariance estimation by CWF takes 1780 seconds for white noise and 2040 seconds for colored noise. The covariance refinement by SDP takes {\color{black} 1.5} seconds, whereas for GS it is {\color{black} less than 1} second. {\color{black}Image denoising by Wiener filtering of 1000 images} takes 84 seconds. For the same images, the runtime for computing contrasts from the Fourier-Bessel coefficients is less than one second which is negligible.

\subsection{Synthetic Data with White Noise}
\begin{figure}[H]
    \centering
    \includegraphics[width=0.9\columnwidth]{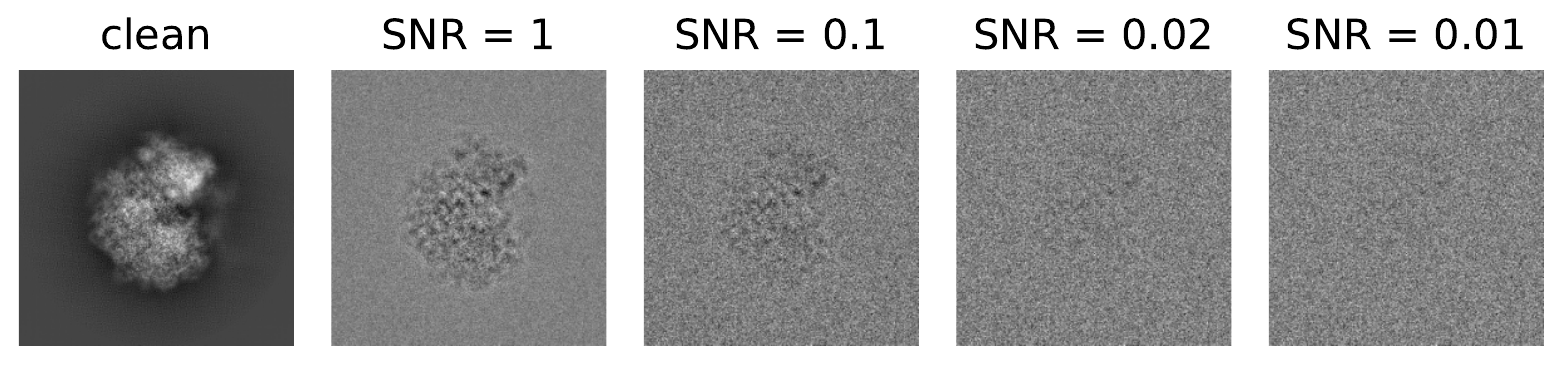}
    \caption{An example of clean and noisy images with white noise. The defocus value for the CTF of the noisy images in this example is 2.67 {\color{black}\si{\mu m}}.}
    \label{fig:noisy_white}
\end{figure}

Figure \ref{fig:noisy_white} shows an example of {\color{black}a} clean image and {\color{black}its noisy counterparts} at different SNRs. 

\begin{figure}[H]
    \centering
    \includegraphics[width=0.5\columnwidth]{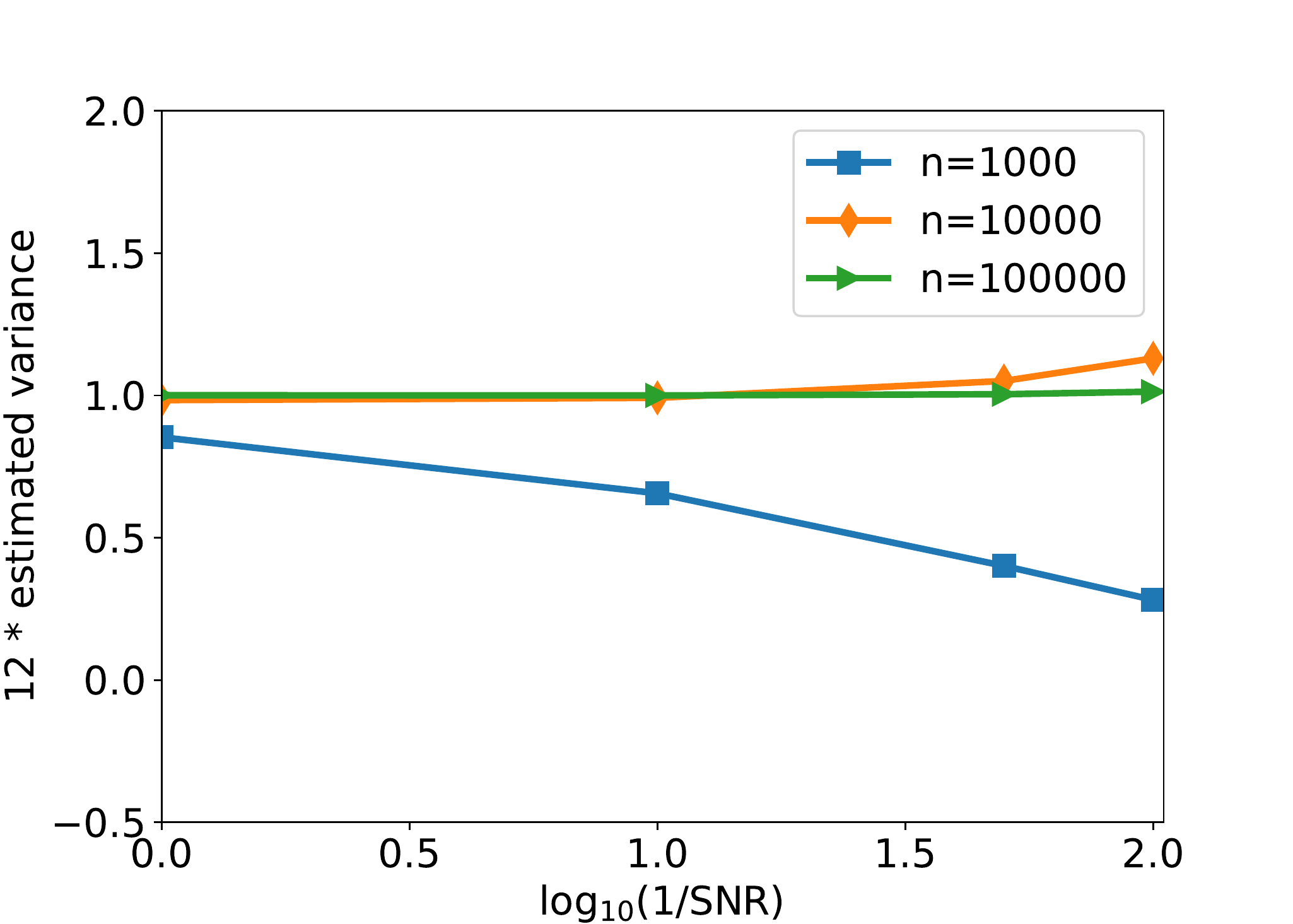}
    \caption{The estimated variance of contrasts with varying SNR and {\color{black}number of images $n$}. {\color{black} The image noise is white Gaussian. The ground truth value of the y-axis is 1, because the image contrasts are sampled from the uniform distribution on $[0.5,1.5]$.}}
    \label{fig:var_white}
\end{figure}

We next examine the performance of our estimator of contrast variance \eqref{eq:var} under {\color{black}different SNRs and number of images.} 
Since our simulated contrasts are uniformly distributed on $[0.5, 1.5]$, the ground truth variance is $1/12$, and thus ideally the line plots in Figure \ref{fig:var_white} should align with the horizontal line  $y=1$. For small number of images $n=1000$, our method often {\color{black}underestimates} the variability of contrast, especially under low SNR. In this regime, our method mainly captures the magnitude of image noise, which is indeed assumed as approximately a constant (so variance is small) across images. For medium size of $n$, namely $n=10000$, our method gives good estimate of contrast variance at SNR$=1, 0.1$, but tends to overestimate it when SNR goes lower. In this regime, the overestimation is mainly due to the inaccurate estimation of $\bmu$ and $\bSig_{cx}$. Ideally, in the absence of noise, $\bmu$ and $\bSig_{cx}\mathbf{1}$ should be parallel to each other due to \eqref{eq:sigma_cx_1}. When $\widehat\bmu$ and $\widehat\bSig_{cx}\mathbf{1}$ are far from being parallel, then one would expect a larger ${\color{black}\Var(c)}$ to minimize the energy in \eqref{eq:var}. We finally remark that when $n=10^5$, we are able to accurately estimate ${\color{black}\Var(c)}$ {\color{black} for SNR as low as $1/100$}.

\begin{figure}[H]
    \centering
    \includegraphics[width=1\columnwidth]{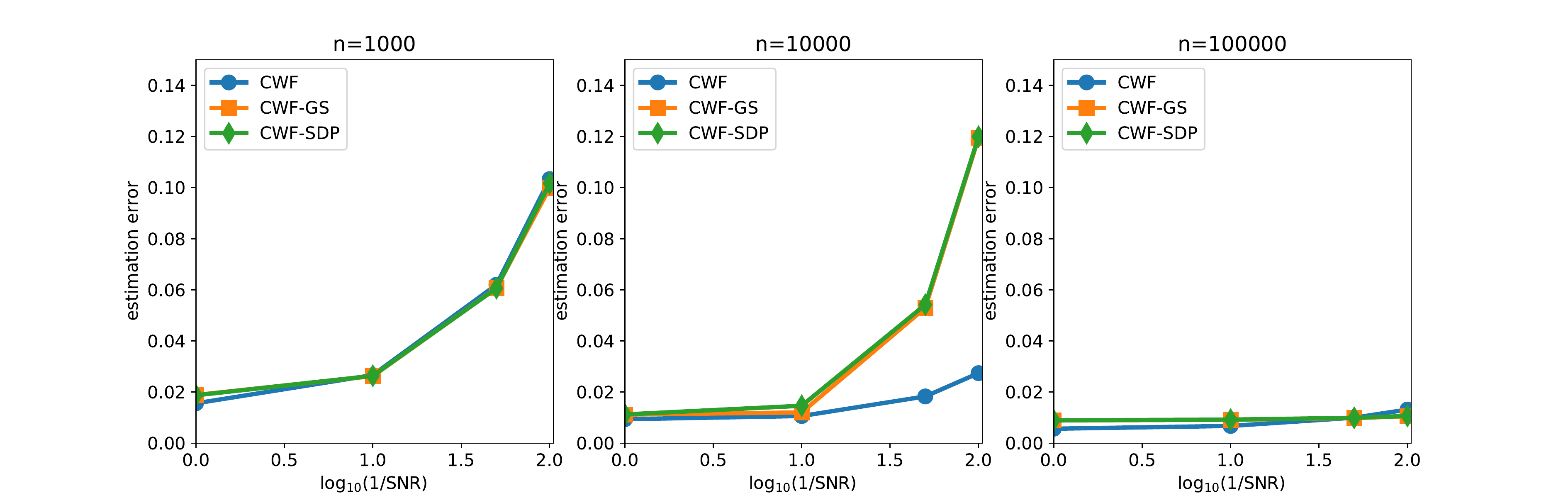}
    \caption{Normalized error of covariance estimates by different methods. {\color{black}The image noise is white Gaussian.}}
    \label{fig:covar_white}
\end{figure}

\begin{figure}[H]
    \centering
    \includegraphics[width=1\columnwidth]{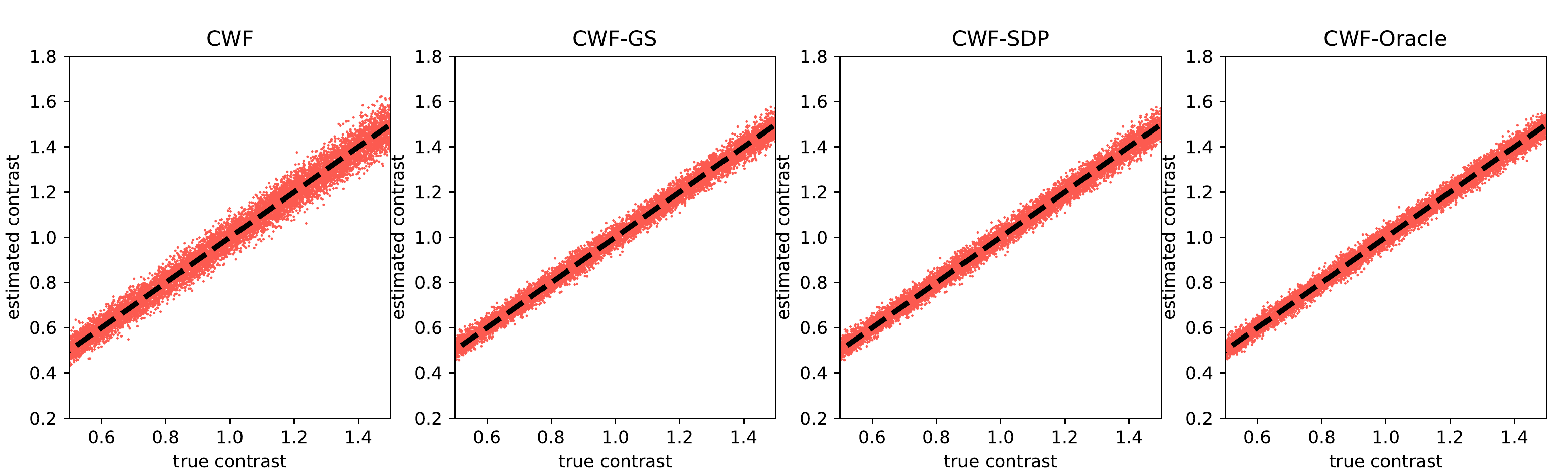}
    \caption{Scatter plots of estimated contrasts v.s. true contrasts. $n=10000$, SNR $=1$. {\color{black} The image noise is white Gaussian. Ideally each scatter plot should align well with the line $y=x$.} }
    \label{fig:scatter_10000_1_white}
\end{figure}

\begin{figure}[H]
    \centering
    \includegraphics[width=1\columnwidth]{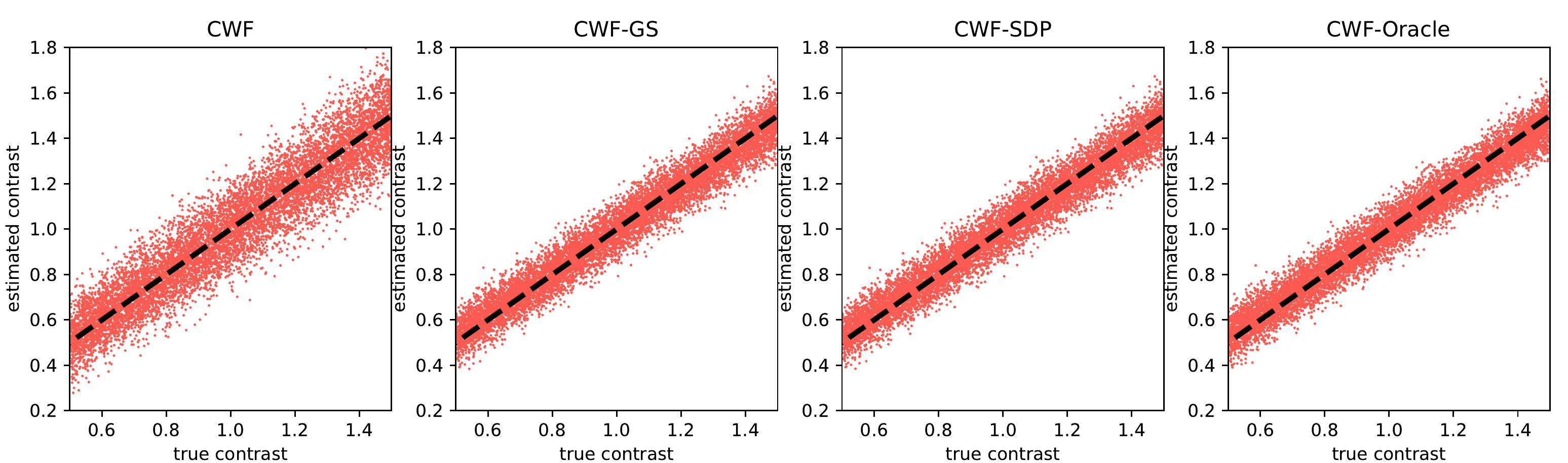}
    \caption{Scatter plots of estimated contrasts v.s. true contrasts. $n=10000$, SNR $=0.1$. {\color{black}The image noise is white Gaussian. Ideally each scatter plot should align well with the line $y=x$.}}
    \label{fig:scatter_10000_01_white}
\end{figure}

\begin{figure}[H]
    \centering
    \includegraphics[width=1\columnwidth]{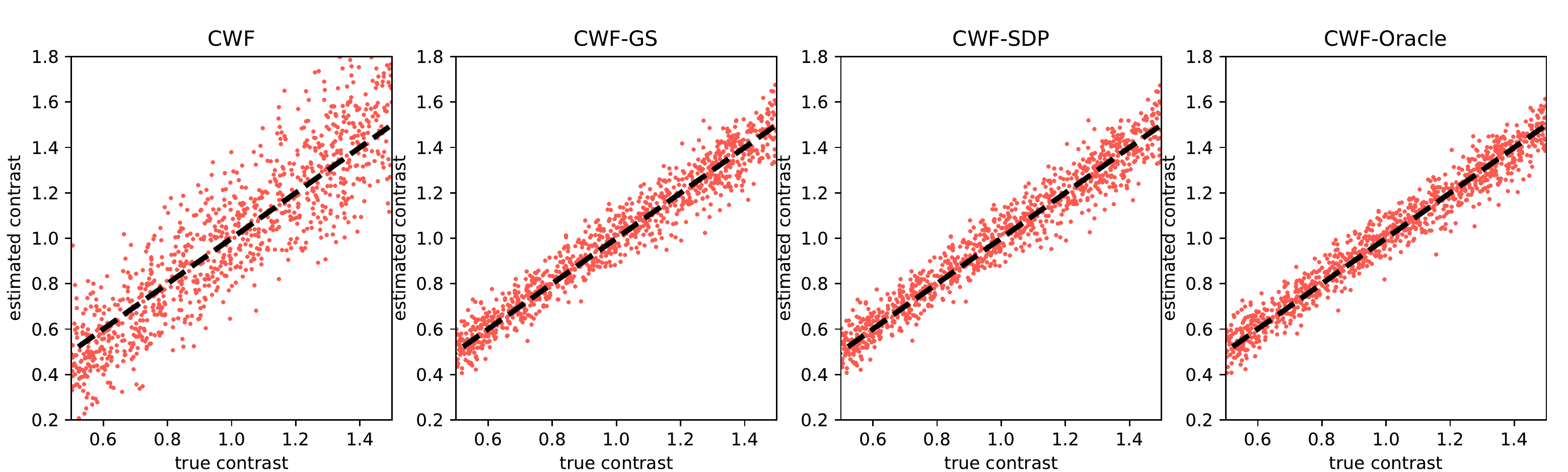}
    \caption{Scatter plots of estimated contrasts v.s. true contrasts. $n=1000$, SNR $=0.1$. {\color{black}The image noise is white Gaussian. Ideally each scatter plot should align well with the line $y=x$.}}
    \label{fig:scatter_1000_01_white}
\end{figure}

We next check the estimation of the covariance matrix $\bSig_{cx}$. In Figure~\ref{fig:covar_white}, we present the line plot of the normalized estimation error $e_{\bSig}:=\|\widehat\bSig_{cx}-\bSig_{cx}\|_F^2/\|\bSig_{cx}\|_F^2$. We observe from Figure \ref{fig:covar_white} that the estimation error of the refined covariance matrix strongly depends on the estimation error of ${\color{black}\Var(c)}$. Indeed, when $n=10000$ and SNR$=1/50$ and $1/100$, $e_{\bSig}$ of CWF-GS and CWF-SDP are both significantly larger than that of CWF. This large error is mainly due to the inaccurate estimation of the contrast variance at those SNRs. Our refined covariance matrices are more accurate under low SNR and large $n$, such as SNR$=1/100$ and $n=10^5$ where ${\color{black}\Var(c)}$ is accurately estimated. We show {\color{black}in Figures \ref{fig:covar_white}-\ref{fig:scatter_1000_01_white}} that although the refinement of covariance matrices does not necessarily reduce the estimation error, it plays a critical role for accurate contrast estimation.

To visualize the quality of contrast estimation by different methods, we present scatter plots of estimated {\color{black}contrasts} v.s. ground truth ones. Ideally, the points in scatter plots should align well with the line $y=x$. We first show in Figure \ref{fig:scatter_10000_1_white} the scatter plots of different methods when $n=10000$ and SNR $=1$.
{\color{black}``CWF-Oracle"} refers to the CWF method with ground truth mean and covariance. We note that the oracle is the best linear estimator of the  contrast.
Our CWF-GS and CWF-SDP perform similarly and both of them achieve near-oracle accuracy for contrast estimation, and they are significantly better than the plain CWF. 

Next, in Figure \ref{fig:scatter_10000_01_white} we keep the number of images {\color{black}fixed} and lower the SNR to 0.1.
All algorithms perform significantly worse than the results of SNR $=1$. However,  CWF-GS and CWF-SDP produce more accurate contrast estimates than those of plain CWF and are comparable to the oracle, which is consistent with Figure \ref{fig:scatter_10000_1_white}.

Next, we show that the number of images often {\color{black}does} not significantly affect the performance of our methods. That is, unlike CWF, our method does not require a large sample size for estimating the contrasts. In Figure \ref{fig:scatter_1000_01_white}, we reduce the number of images to 1000 while keeping SNR $=0.1$.
The contrast estimation by the plain CWF is much less accurate after reducing the number of images. In contrast, our methods better maintain the quality of contrast estimates after reducing $n$. This suggests that our method is more robust to inaccuracies of the estimated covariance matrix. 

\begin{figure}[H]
    \centering
    \includegraphics[width=1\columnwidth]{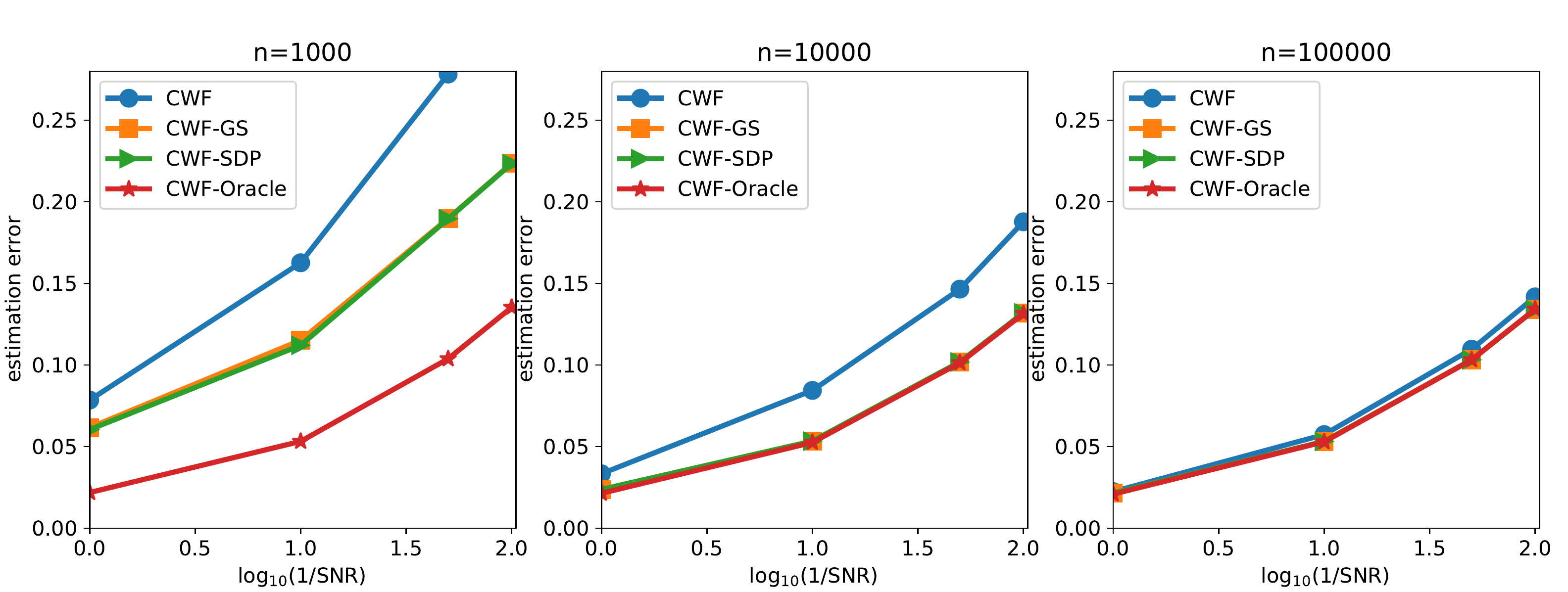}
    \caption{Contrast estimation error under different SNRs and  number of images. {\color{black} The image noise is white Gaussian.}}
    \label{fig:line_256_white}
\end{figure}

In Figure \ref{fig:line_256_white} we compare the contrast estimation error of different methods under different SNRs and number of images.
We use the averaged relative error 
\begin{align}
    e_c = \frac{1}{n}\sum_{i=1}^n\left|\frac{\widehat c_i - c_i}{c_i}\right|
\end{align}
to measure the performance of the contrast estimation. We limit the y-axis of the line plot on the interval $[0, 0.28]$ since any contrast estimation error above 0.28 {\color{black}is} regarded as non-informative. Indeed, a trivial contrast estimator that estimates every $c_i$ as 1 would give the error close to 0.28 in expectation.
We observe that when $n=10000$, although the covariance matrices are not very accurately estimated, CWF-GS and CWF-SDP both achieve performance that is comparable to the oracle. However, CWF needed  $100000$ samples to reduce the gap to the oracle. Even {\color{black}with} $n=100000$, CWF is still slightly worse than the oracle and our methods. Therefore, the key factor that determines the quality of contrast estimation is not how covariance is close to the true one, but is whether the covariance is enforced to satisfy  {\color{black}the constraints stated in} Proposition \ref{prop:constr}.

{\color{black}Next}, we test the performance of the algorithms on image denoising. We compare the plain CWF and the ones with our refined covariances. We also compare the denoised {\color{black}images} with image normalization and our 2-stage CWF procedure, introduced in Section \ref{sec:denoise}. The two previous methods we compare are CWF and CWF-norm \cite{CWF}. The latter one is the CWF with an image normalization step. The labels ``-GS" and ``-SDP" refer to usage of the refined covariance matrix (estimated by our GS procedure and SDP method) for CWF.

\begin{figure}[H]
    \centering
    \includegraphics[width=1\columnwidth]{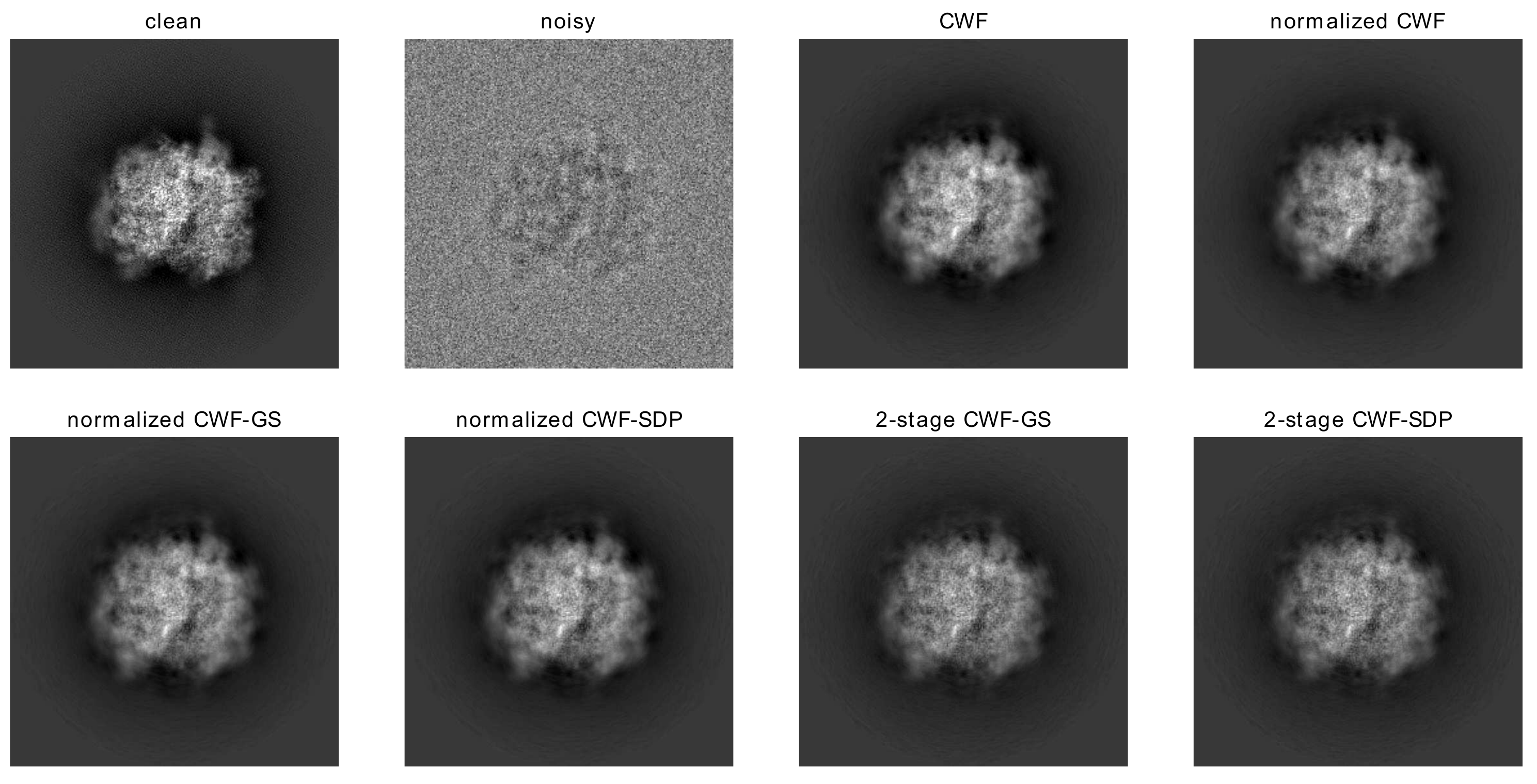}
    \caption{Clean, noisy and denoised images with SNR $=0.1$ and $n=10000$. {\color{black} The image noise is white Gaussian.}}
    \label{fig:denoise_10000_01_white}
\end{figure}

Before presenting the estimation errors, we show an example of clean and noisy images and denoised ones by different methods. In this example, SNR $=0.1$ and $n=10000$.
From the result of Figure \ref{fig:denoise_10000_01_white}, the denoised image by the original CWF with normalization looks similar to the ones by our normalization methods, although they have slightly different contrasts. The denoised images by our 2-stage methods have clearer fine details than those that are denoised by other methods.

\begin{figure}[H]
    \centering
    \includegraphics[width=1\columnwidth]{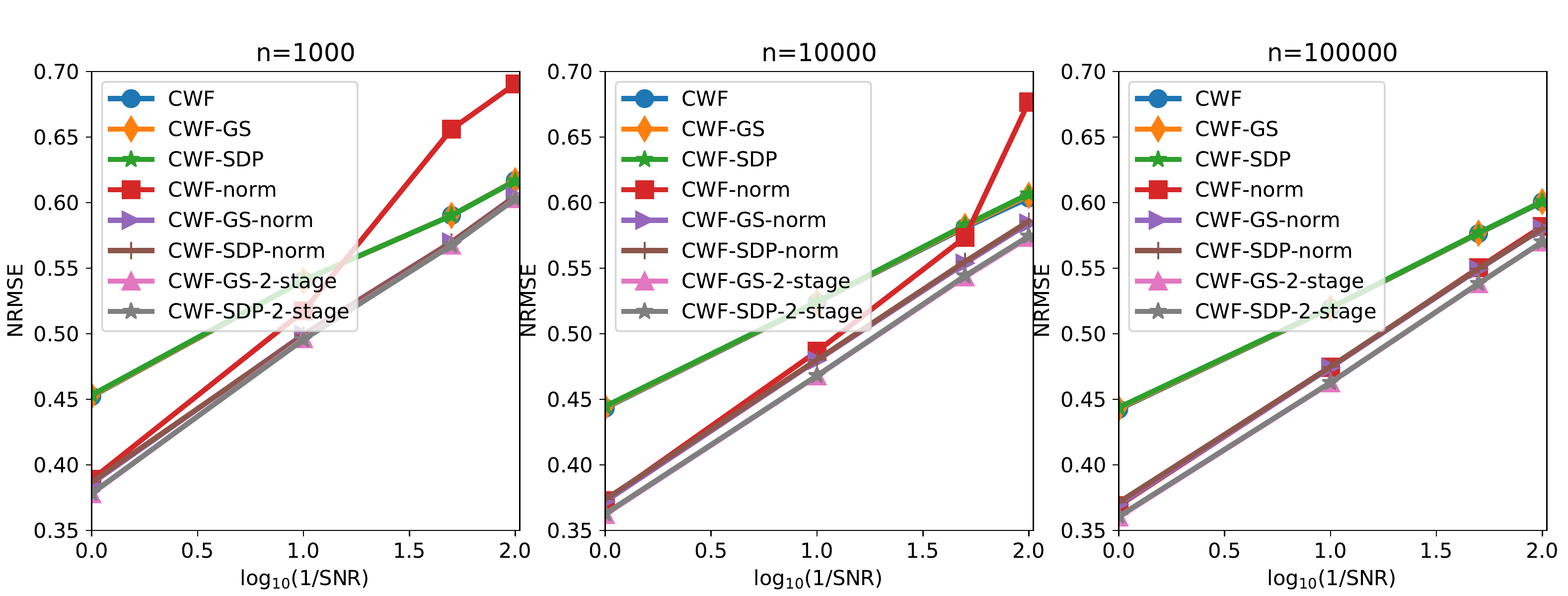}
    \caption{NRMSE of the denoised images under different SNRs and the number of images. {\color{black} The image noise is white Gaussian.}}
    \label{fig:denoise_line_white}
\end{figure}

We evaluate the denoising performance by the normalized root mean squared error (NRMSE) {\color{black} within a circular mask whose radius is half the image size.}
From Figure \ref{fig:denoise_line_white}, CWF with image normalization often gives large {\color{black}errors} under low SNR and small to medium $n$. Our image normalization and 2-staged methods consistently perform better than CWF and CWF-normalization, where 2-staged methods are slightly better. We also observe that our GS and SDP refinement yield similar estimation errors, where GS is slightly better under low SNRs.

{\color{black}
We further examine the contrast estimation error in each defocus group. In Figure \ref{fig:contrast_white_df} we compare the contrast estimation errors of different methods in each of the 10 defocus groups for $n=10000$ and SNR $=0.1$. The defocus groups are sorted by defocus values in ascending order. In addition to the previously tested methods, we include a stronger oracle that knows the true clean images (not just true covariance). It estimates the contrast by 
   $ \widehat c_i = \langle\by_i\,, A_i\bx_i\rangle/\|A_i\bx_i\|^2_2$. We refer to this method as ``Oracle". On the right panel of Figure \ref{fig:contrast_white_df} we test the contrast estimation when the observed noisy images are randomly shifted by 1-5 pixels in x and y directions.
   From Figure \ref{fig:contrast_white_df}, the contrast estimation errors of both CWF and our methods tend to decrease when the defocus value increases. This makes sense, since CTFs with larger defocus values have higher absolute values around the zero-th frequency, and thus enjoy higher SNRs at low frequencies. When all images are centered, the ``oracle" indicates the best possible contrast estimation that a template-based method can achieve, which obviously outperforms all other methods including the CWF-oracle. We remark that ``oracle" knows the true manifold of the clean images, whereas ``CWF-oracle"  assumes a linear approximation of it. However, the new oracle is not robust to shifts, unlike other methods. When the noisy images are shifted, the oracle, assuming it does not know the shifts in the observation and only computes the dot product between the shifted $\by_i$ and centered $A_i\bx_i$, gives poor contrast estimation. To mitigate this issue, a low pass filter to $\by_i$ and $A_i\bx_i$ is often needed.
\begin{figure}[H]
    \centering
    \includegraphics[width=1\columnwidth]{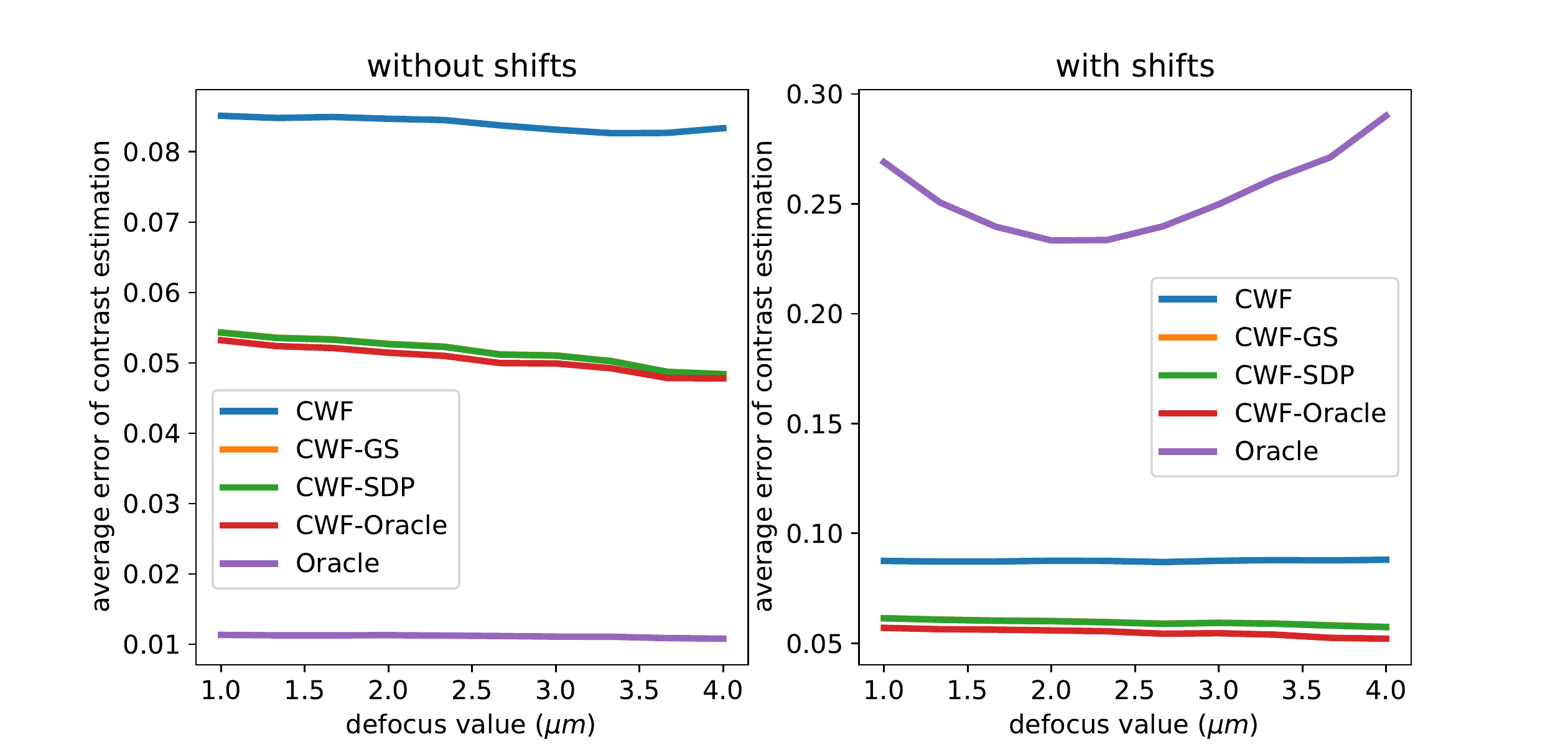}
    \caption{{\color{black}Average error per defocus group of contrast estimation by different methods. $n=10000$, SNR $=0.1$. The image noise is white Gaussian. The left figure panel uses centered noisy images. In the right panel, we randomly shifted the noisy images by 1-5 pixels in the x and y directions independently. In both panels, the two lines corresponding to CWF-GS and CWF-SDP overlap with each other.}}
    \label{fig:contrast_white_df}
\end{figure}

At last, in the left panel of Figure \ref{fig:err_white_df}, we compare the NRMSE of the denoised images by CWF with image normalization and our methods for each defocus group. On the right panel we show the relationship between the NRMSE of the denoised images and their contrast values. In particular, we  divide the images into 10 groups by their true contrast values. Namely, the images with contrasts between 0.5 and 0.6 are classified as the first contrast group, and those with contrasts 0.6-0.7 are considered the second group and so on. We do not show CWF with image normalization since it has significantly higher NRMSE than other methods and will screw the scale of the y-axis.
From the figure, for all methods, the NSMSEs often decrease when defocus values and contrast values increase. This agrees with our argument that higher defocus and contrast correspond to higher SNRs at low frequencies. However, with higher defocus values, more energy of clean signals is spilled outside of the image disk \cite{CTF_SINDELAR}, and CTFs have more zero-crossings, which may have negative effects on image denoising. Indeed, we notice that when defocus values approach 4 {\color{black}\si{\mu m}}, the NRMSEs slightly increase. Overall, the 2-stage methods perform significantly better than other methods.
\begin{figure}[H]
    \centering
    \includegraphics[width=1\columnwidth]{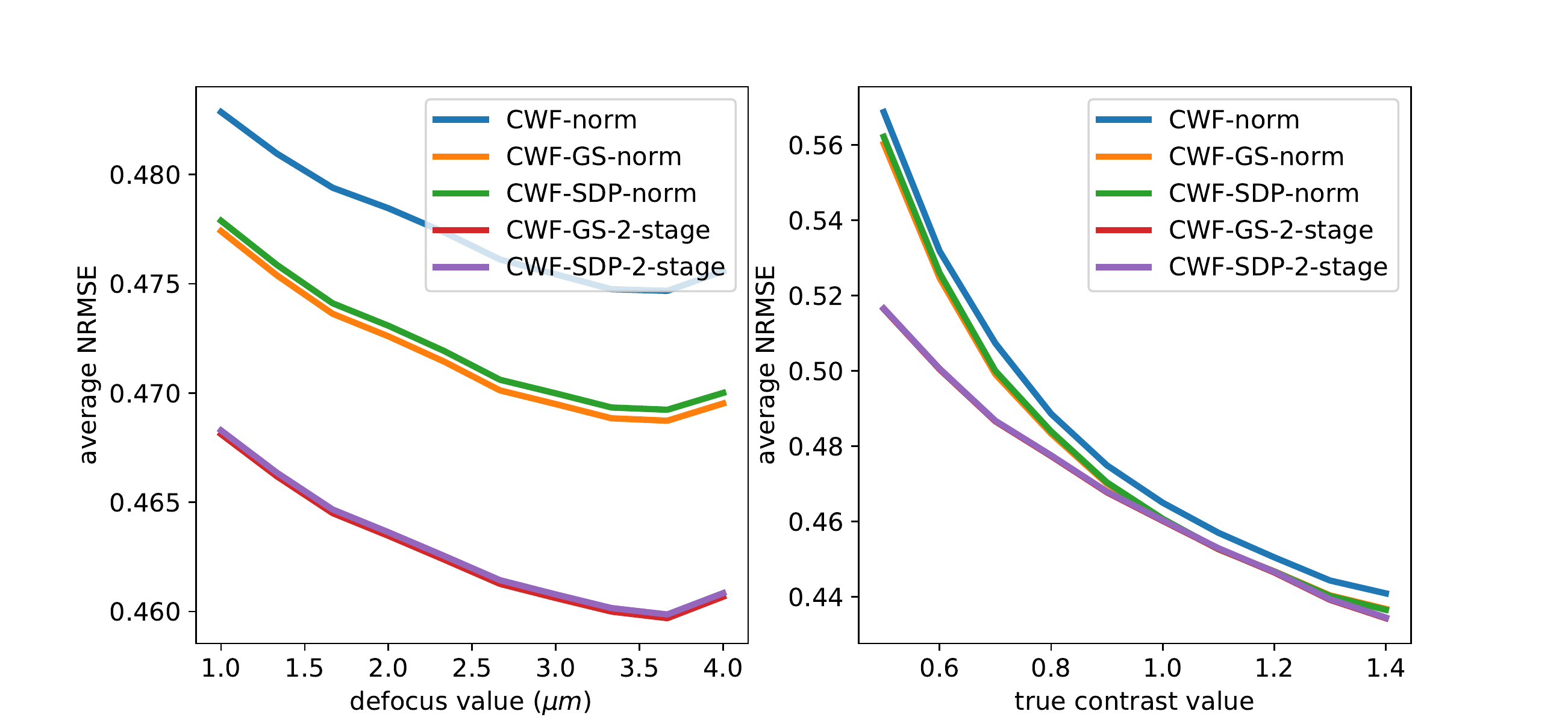}
    \caption{{\color{black}Average NRMSE of denoised images from different methods, per defocus group (left figure) and per contrast group (right figure). $n=10000$, SNR $=0.1$. The image noise is white Gaussian. In the right panel, the red and purple lines overlap with each other.}}
    \label{fig:err_white_df}
\end{figure}

}
\subsection{Synthetic Data with Colored Noise}
We retest different methods on synthetic data with colored noise. The data generation procedure is exactly the same as before, except that now we use colored noise whose power spectrum decays with the radial frequency. {\color{black}Colored noise is more realistic in the sense that it better mimics the noise statistics observed in experimental images. Our choice of colored noise makes contrast estimation more challenging.} Indeed, given the noise spectrum $1/\sqrt{k^2+1}$ (up to a constant) {\color{black} where $k$ is measured in 1/(128 pixel size)}, under the same SNR, the noise power spectrum in the zeroth frequency is expected to be 40 times larger than that of the white noise. Since contrast (mean of the pixels) is all about the zeroth frequency, the high noise {\color{black}at low frequencies}  poses a serious challenge. 

\begin{figure}[H]
    \centering
    \includegraphics[width=0.9\columnwidth]{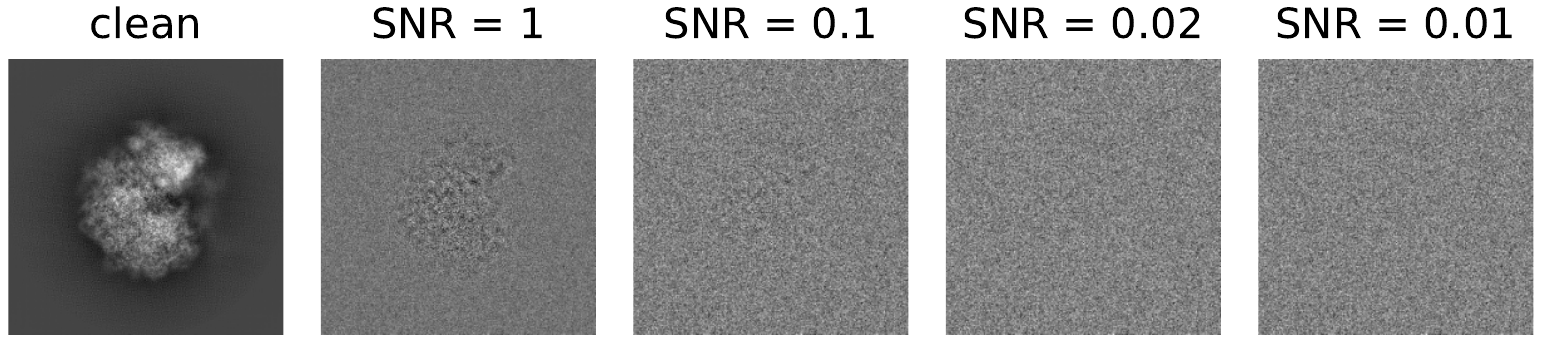}
    \caption{An example of clean and noisy images with colored noise. The defocus value for the CTF of the noisy images in this example is 2.67 {\color{black}\si{\mu m}}.}
    \label{fig:noisy_color}
\end{figure}

Figure \ref{fig:noisy_color} shows an example of a clean image and noisy ones at different SNRs.
Comparing with Figure \ref{fig:noisy_white}, the particles are harder to identify by human eyes than in the case of white noise. Indeed, starting from SNR=0.1, it already becomes hard to visually distinguish the particle from the colored noise in the background.

\begin{figure}[H]
    \centering
    \includegraphics[width=0.6\columnwidth]{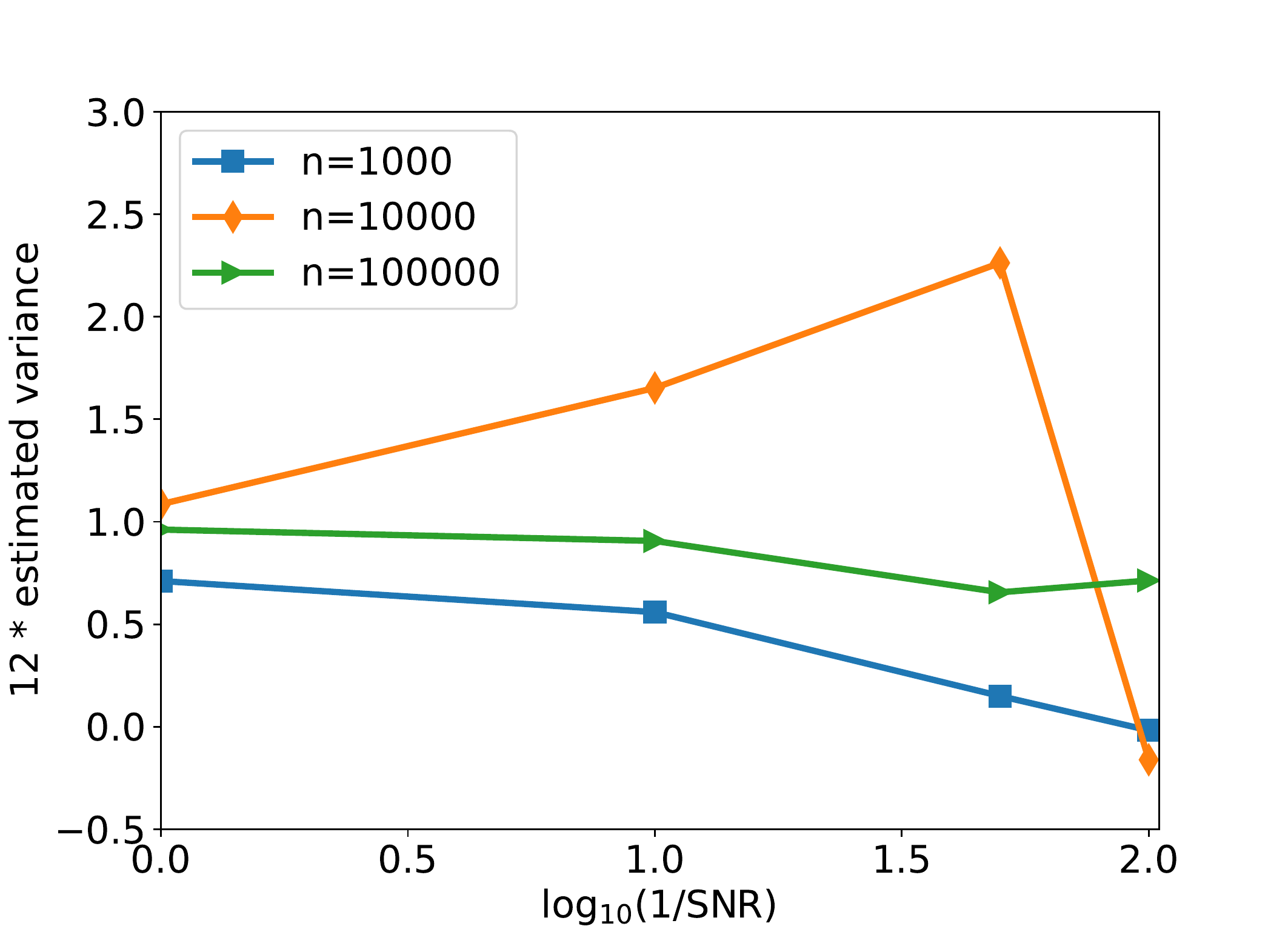}
    \caption{The estimated variance of contrasts with varying SNR and $n$. {\color{black}The image noise is colored Gaussian with decaying PSD. The ground truth of the y-axis is 1.}}
    \label{fig:var_color}
\end{figure}

We next test the variance estimation for the contrasts. 
As shown in Figure \ref{fig:var_color}, the performance of the variance estimation is indeed worse than that for images with white noise. For $n=1000$, our method consistently {\color{black}underestimates} the variance. For $n=10000$, there is an interesting transition from overestimation to underestimation between SNR$=1/50$ and $1/100$. This is likely due to that at SNR$=1/100$, our method starts to learn the variance of the average pixel values of noise which is close to 0. However, for $n=100000$, we are able to reliably estimate ${\color{black}\Var(c)}$ up to SNR $=0.1$.

\begin{figure}[H]
    \centering
    \includegraphics[width=1\columnwidth]{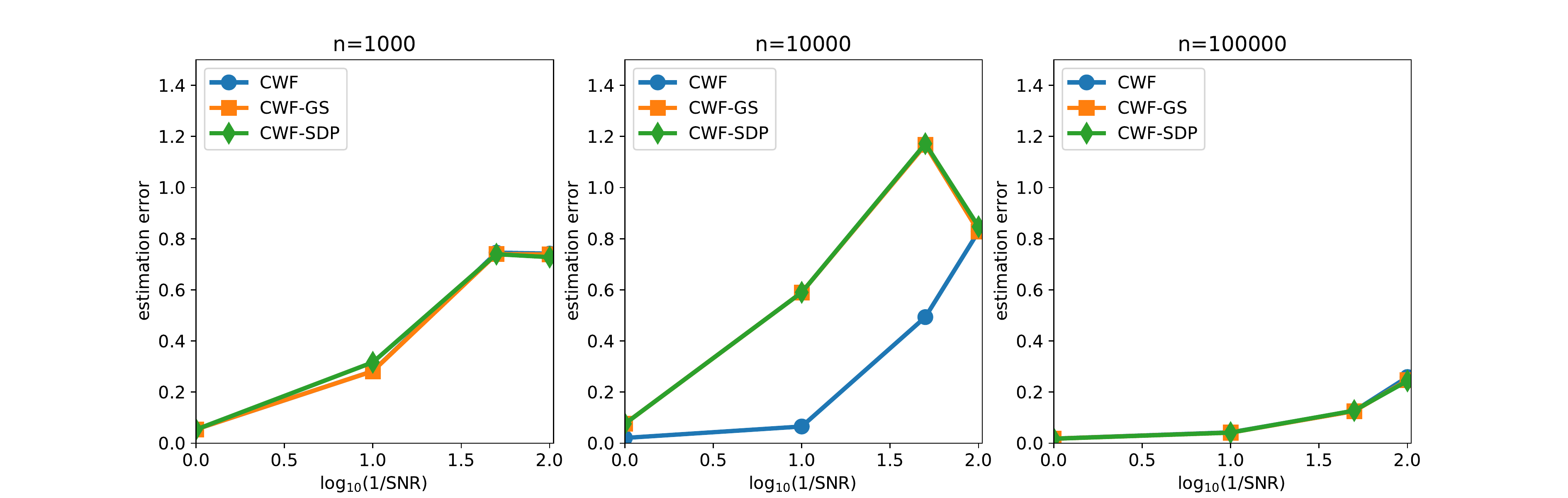}
    \caption{Normalized error of covariance estimates by different methods. {\color{black}The image noise is colored Gaussian with decaying PSD. The line of CWF does not appear in the left panel due to its high error. In the right panel, the line of CWF overlaps with the lines of other methods.}}
    \label{fig:covar_color}
\end{figure}

Figure \ref{fig:covar_color} shows the covariance estimation error for the different methods.  
Similar to the white noise case, the large errors of our methods when $n=10000$ are mainly due to overestimation of ${\color{black}\Var(c)}$. {\color{black}We notice a slight drop of covariance estimation error at SNR$=0.01$ when $n=10000$. This is due to the reduced error of contrast variance estimation (see the orange line in Figure \ref{fig:var_color}).} However, as we show next, these refined covariance matrices are key for accurate contrast estimation.

\begin{figure}[H]
    \centering
    \includegraphics[width=1\columnwidth]{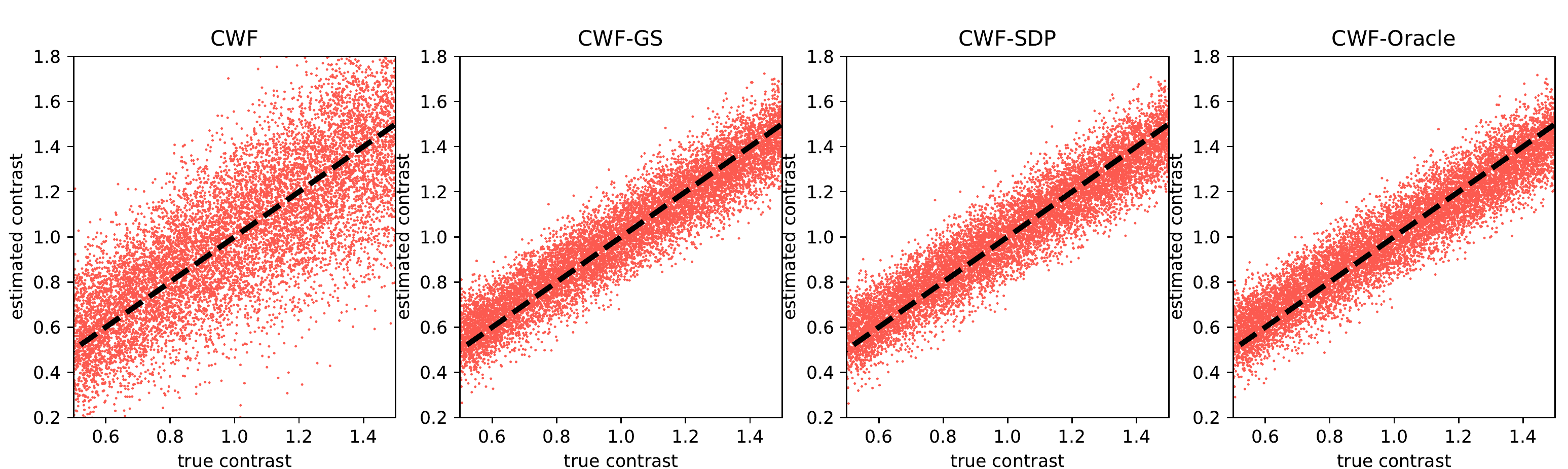}
    \caption{Scatter plots of estimated contrasts v.s. true contrasts. $n=10000$, SNR $=1$. {\color{black}The image noise is colored Gaussian with decaying PSD.}}
    \label{fig:scatter_10000_1_color}
\end{figure}

As before, we assess the quality of  the contrast estimation through scatter plots.
From the result of Figure \ref{fig:scatter_10000_1_color}, all methods perform significantly worse than in the white noise case. However our methods are still comparable to the oracle one and are considerably better than the original CWF. 

\begin{figure}[H]
    \centering
    \includegraphics[width=1\columnwidth]{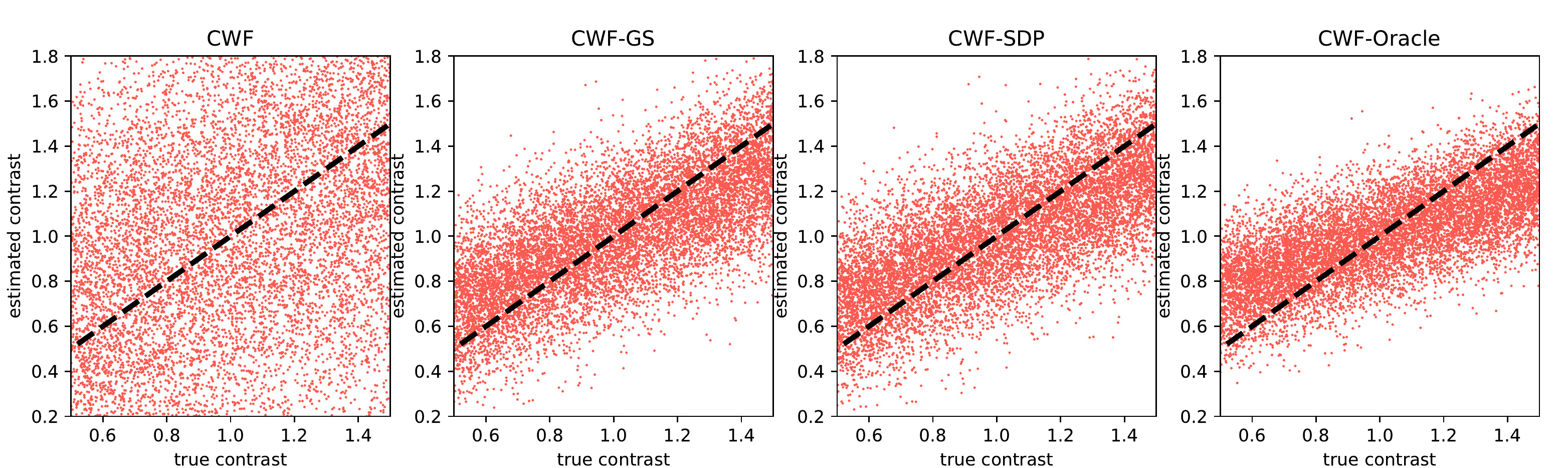}
    \caption{Scatter plots of estimated contrasts v.s. true contrasts. $n=10000$, SNR $=0.1$. {\color{black}The image noise is colored Gaussian with decaying PSD.}}
    \label{fig:scatter_10000_01_color}
\end{figure}

Next, we fix $n$ and decrease SNR to 0.1.
In Figure \ref{fig:scatter_10000_01_color}, the original CWF almost fails since there is no clear linear association between its estimated contrasts and the true ones. However, one can see a clear trend between the contrasts estimated by our methods and the ground truth ones. Again, our methods achieve comparable accuracy to the oracle one.

\begin{figure}[H]
    \centering
    \includegraphics[width=1\columnwidth]{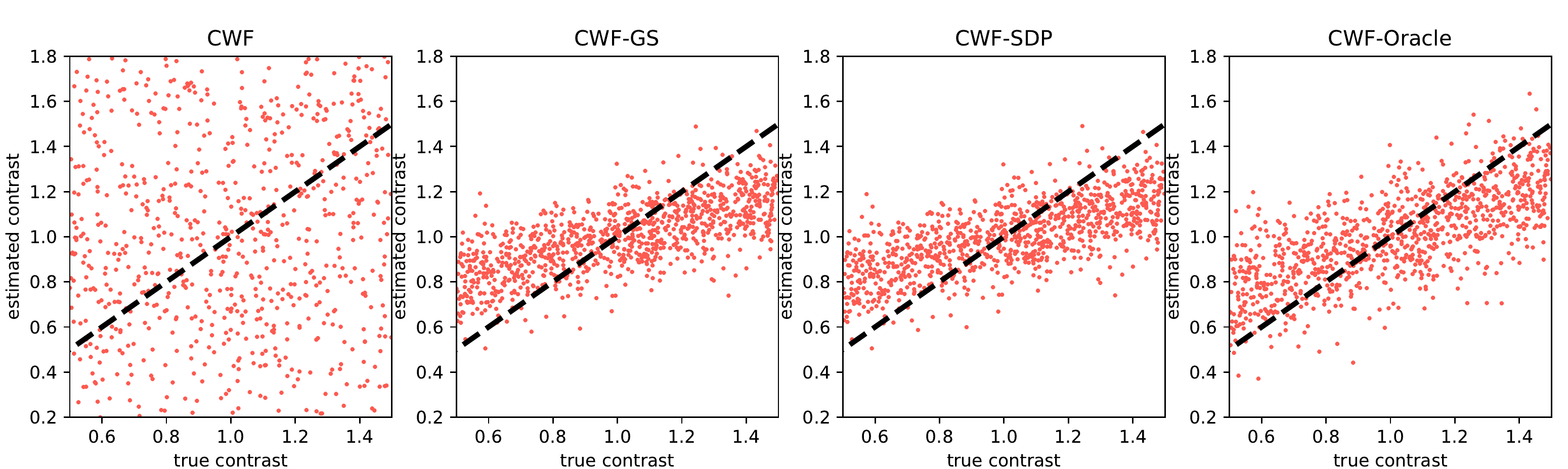}
    \caption{Scatter plots of estimated contrasts v.s. true contrasts. $n=1000$, SNR $=0.1$. {\color{black}The image noise is colored Gaussian with decaying PSD.}}
    \label{fig:scatter_1000_01_color}
\end{figure}

We next keep the SNR and reduce the number of images to 1000.
In Figure \ref{fig:scatter_1000_01_color}, the performance gap between our methods and CWF is even larger, and our methods are still comparable to the oracle. {\color{black}In both Figure \ref{fig:scatter_10000_01_color} and \ref{fig:scatter_1000_01_color} the scatter plots of our methods tend to follow a straight line with a smaller slope, due to the high noise. Indeed, consider the extreme case of pure noise images, the estimated contrasts should follow a horizontal line. }

\begin{figure}[H]
    \centering
    \includegraphics[width=1\columnwidth]{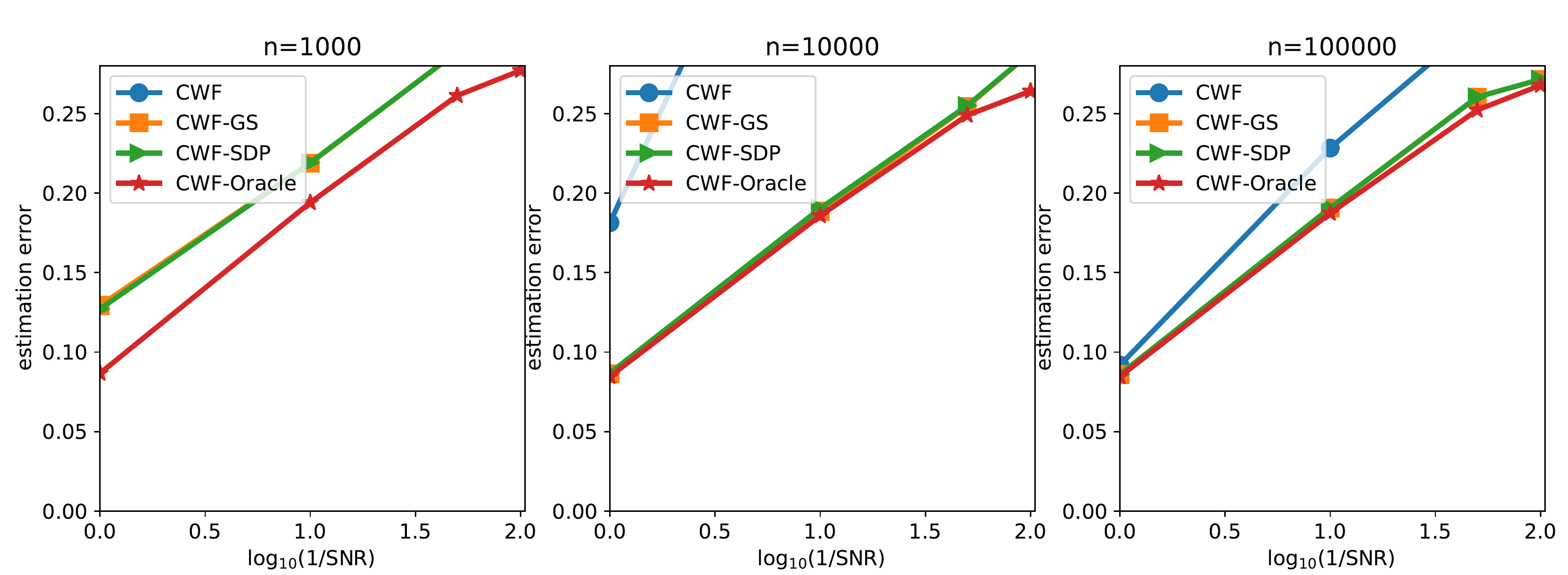}
    \caption{Contrast estimation error under different SNRs and  number of images. {\color{black}The image noise is colored Gaussian with decaying PSD.}}
    \label{fig:line_256_color}
\end{figure}

In Figure \ref{fig:line_256_color} we compare the contrast estimation error of different methods under different SNRs and number of images.
We observe that the CWF-GS and CWF-SDP both perform comparably to the oracle for $n\geq 10000$. They also perform  close to the oracle for the small sample {\color{black}size $n=1000$}, which indicates their robustness to the sample size unlike CWF. When $n=1000$, the error of CWF is {\color{black}always} above 0.28, {\color{black} thus it} does not appear in the plot. There is still a large gap between CWF and {\color{black}our methods} (and oracle) when $n=100000$. 

\begin{figure}[!htbp]
    \centering
    \includegraphics[width=1\columnwidth]{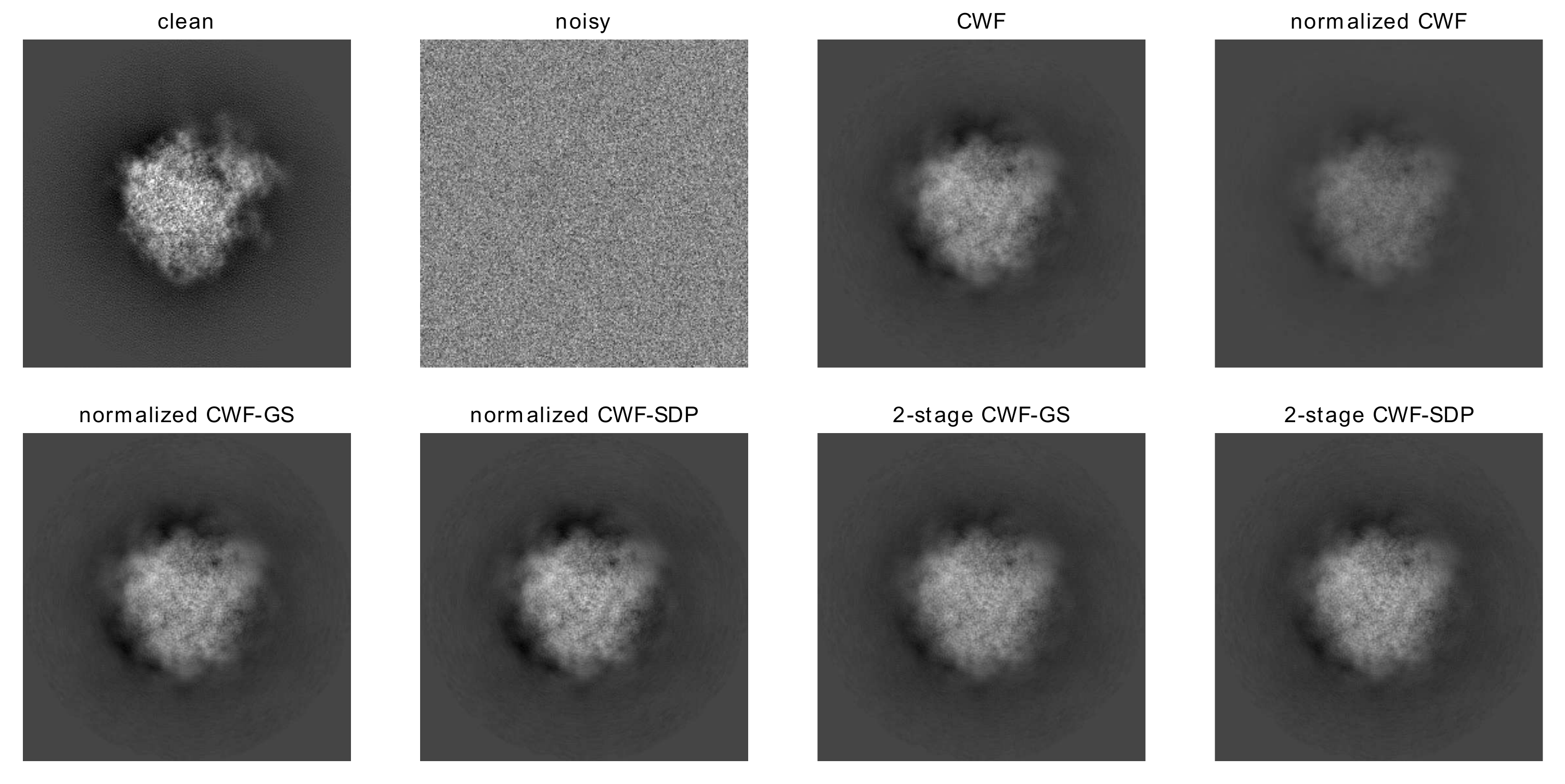}
    \caption{Clean, noisy and denoised images with SNR $=0.1$ and $n=10000$. {\color{black}The image noise is colored Gaussian with decaying PSD.}}
    \label{fig:denoise_10000_01_color}
\end{figure}

As for image denoising,
we first show an example of clean and noisy images and denoised ones by different methods. In this example, SNR $=0.1$ and $n=10000$.
From the result of Figure \ref{fig:denoise_10000_01_color}, the denoised image by the original CWF with normalization gives much lower contrast than  the ones by our normalization methods. The denoised images by our 2-stage methods seem to have clearer fine details than those that are denoised by other methods.

\begin{figure}[H]
    \centering
    \includegraphics[width=0.9\columnwidth]{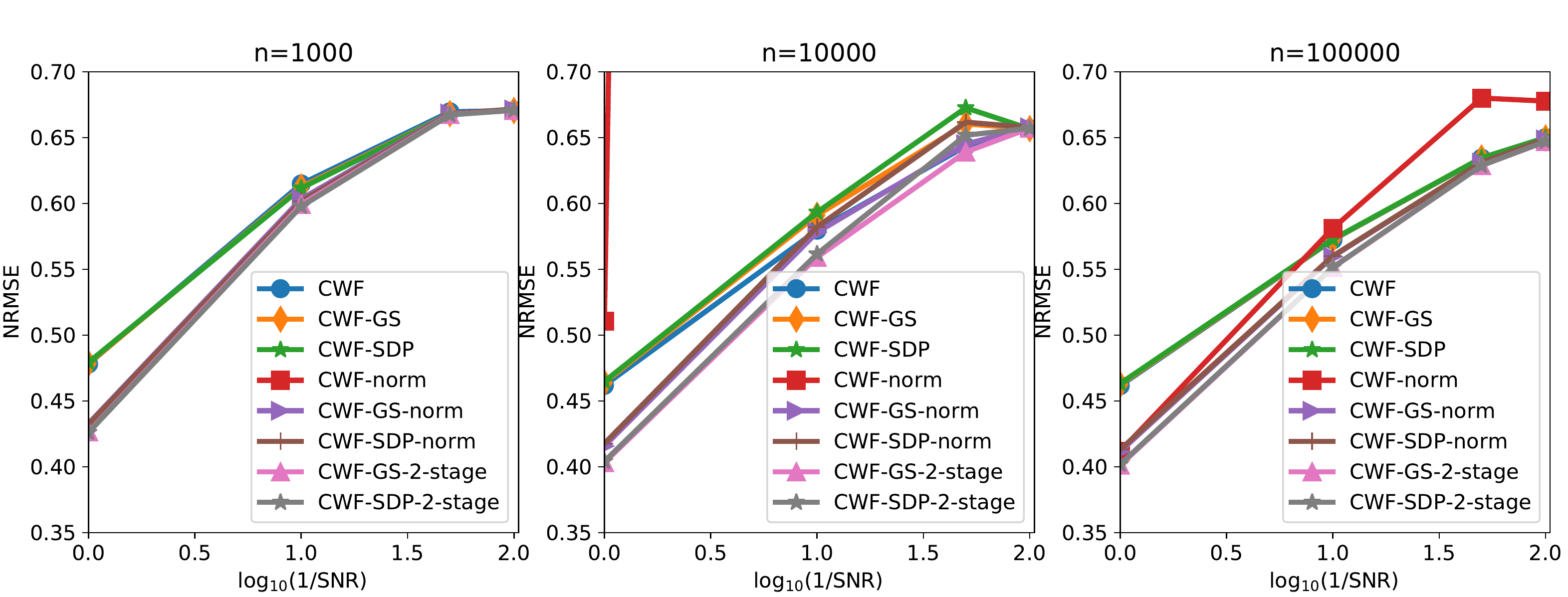}
    \caption{NRMSE of the denoised images under different SNRs and number of images. {\color{black}The image noise is colored Gaussian with decaying PSD.}}
    \label{fig:denoise_line_color}
\end{figure}

{\color{black}Next}, we compare the NRMSE of the denoised images by the different algorithms. 
From Figure \ref{fig:denoise_line_color}, CWF with image normalization is very unstable. It does not appear in the first subplot due to exceeding the y-axis limit. Similar to the white noise case, our image normalization and 2-staged methods often have smaller errors than other methods, where 2-staged methods are slightly better. Similar to the white noise case, our GS refinement yields slightly smaller estimation errors than the SDP method under low SNRs.

{\color{black}Similar to the white noise case, we examine the relationship between contrast estimation errors and the defocus values of the corresponding CTFs. In Figure \ref{fig:contrast_color_df} we compare the average contrast estimation errors of different methods in each of the 10 defocus groups. The defocus groups are sorted by defocus values in ascending order. On the right of Figure \ref{fig:contrast_color_df} we test the contrast estimation when the observed noisy images are randomly shifted by 1-5 pixels in the x and y directions.
From Figure \ref{fig:contrast_color_df}, the contrast estimation errors of both CWF and our methods tend to decrease when defocus value increases. The instability of the ``oracle" method to the shifts of images is also observed. 
\begin{figure}[H]
    \centering
    \includegraphics[width=1\columnwidth]{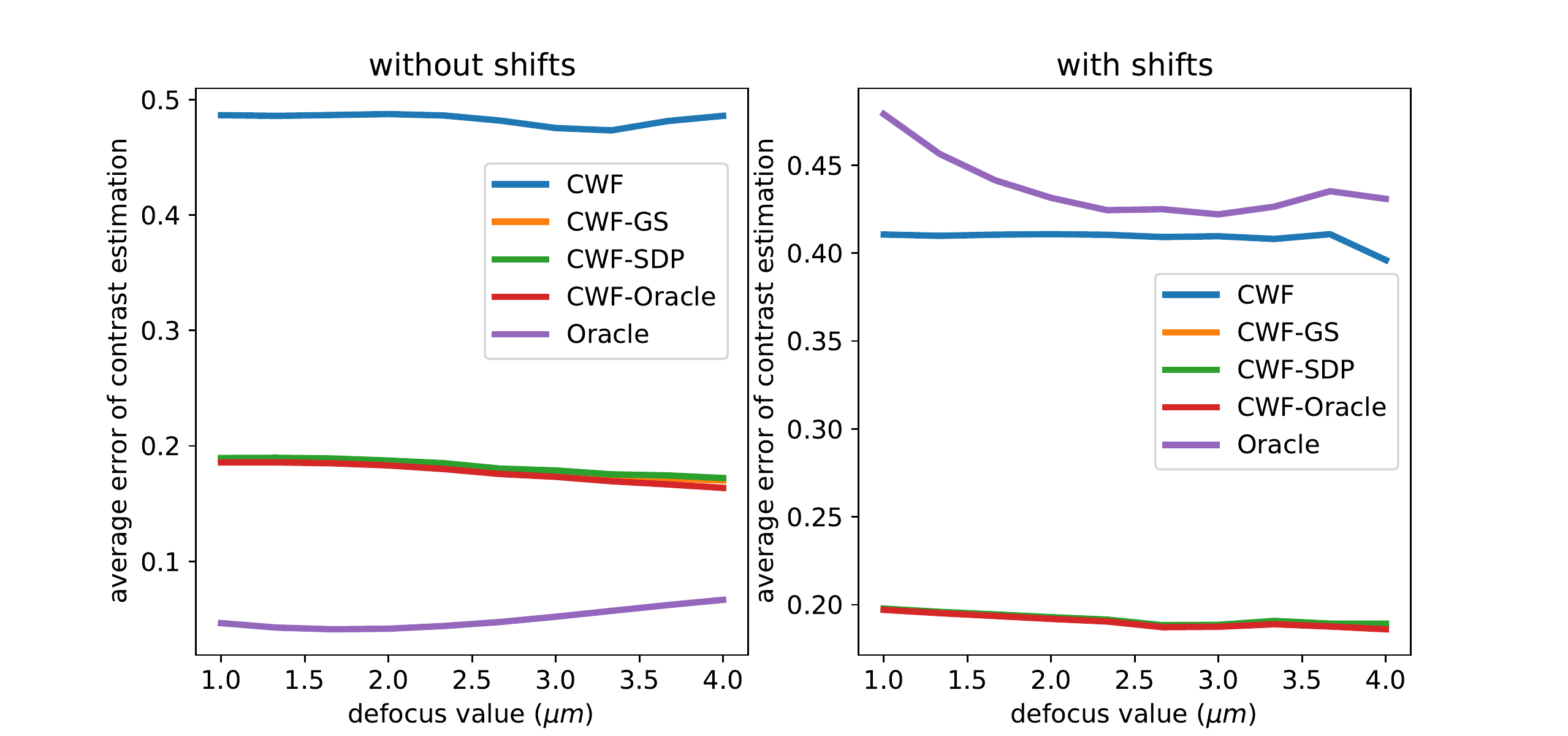}
    \caption{{\color{black}Average error per defocus group of contrast estimation by different methods. $n=10000$, SNR $=0.1$. The image noise is colored Gaussian with decaying PSD. The left panel uses centered noisy images. In the right panel, we randomly shift noisy images by 1-5 pixels in the x and y directions independently. In the right panel, the lines corresponding to CWF-GS and CWF-SDP overlap with each other.}}
    \label{fig:contrast_color_df}
\end{figure}

\begin{figure}[H]
    \centering
    \includegraphics[width=1\columnwidth]{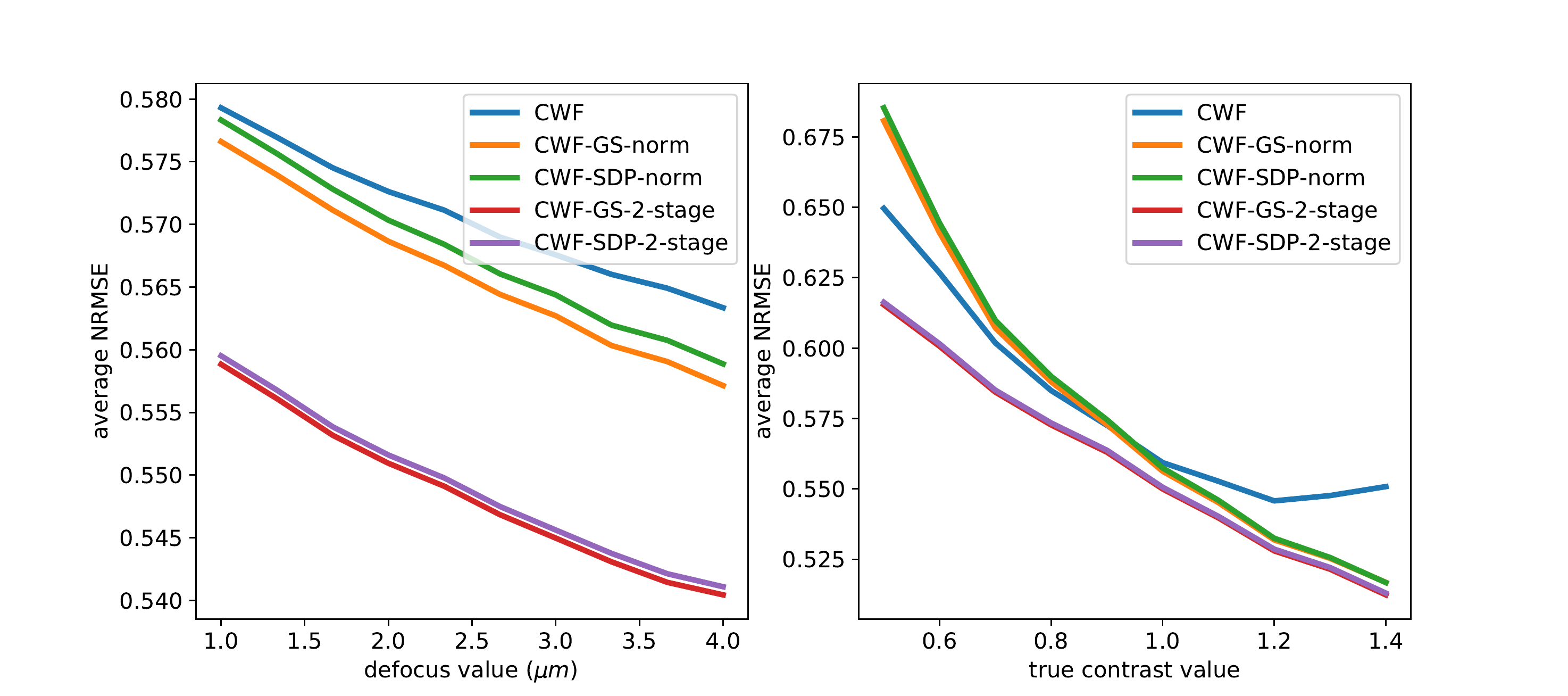}
    \caption{{\color{black}Average NRMSE of denoised images from different methods, per defocus group (left panel) and per contrast group (right panel). $n=10000$, SNR $=0.1$. The image noise is colored Gaussian with decaying PSD.}}
    \label{fig:err_color_df}
\end{figure}

Same as the white noise case, in the left panel of Figure \ref{fig:err_color_df}, we compare the NRMSE of the denoised images by CWF and our methods within each defocus group. The right panel shows the relationship between NRMSE of denoised images and their contrast values. 
From the figure, for all methods, the NSMSE often decreases when defocus value and contrast value increase. The results of this section suggest that the 2-stage methods outperform the other CWF-based methods, and we therefore expect them to be the method of choice also for experimental data.

}

\section{Results for Experimental Data}
We compare our methods with CWF on three experimental datasets, which are freely downloadable from the Electron Microscope Pilot Image Archive (EMPIAR) database \cite{empiar}. {\color{black}We chose these datasets for a purely technical reason, as each micrograph in these datasets has a single CTF, which reduces the total number of CTFs and runtime of our method.   For the datasets where each image has its own CTF, it is  possible accelerate our method by implementing in 2-D Fourier space the operations that involve the CTFs. However, to keep the idea of this work clean and focused, we leave this modification to future work. Due to the similar performance of our methods on the the three datasets, in this section we only present the result for EMPIAR-10028 \cite{10028}, and refer the reader to the appendix for the results for EMPIAR-10005 \cite{10005} and EMPIAR-10073 \cite{10073}.}

For all datasets, {\color{black}we first normalize each individual image by the standard deviation (std) of the pixel values at the image corners  that are located outside a circular mask with radius $0.45L$, where $L$ is the dimension of the square image. Next, for each defocus group we estimate the PSD of the noise in the normalized images, using the pixel values outside of the same mask. We then perform background subtraction by subtracting the mean of pixel values outside the mask.} For each defocus group, the images are whitened by applying the single whitening filter that equals to the $-0.5$ power of the estimated noise PSD of that defocus group.
By doing this, we are assuming that the images {\color{black}in the same micrograph have similar noise PSDs, in order to reduce the estimation error of the noise PSD which could be quite large for a single image. Moreover, whitening by defocus group accelerates our algorithm. In particular, given the whitened image formation model $\bW_i\by_i = c_i\bW_i\bA_i\bx_i + \bW_i\beps_i$, to recover $\bx_i$ we use the whitened CTFs $\bW_i\bA_i$. Using distinct whitening filter $\bW_i$ for all images increases the number of distinct CTFs and the computational complexity of the existing implementation of CWF. We also note that we have to estimate the PSD before the background subtraction, otherwise the estimated PSD will vanish at the zeroth frequency and cause numerical issues when whitening the image.}
The possibly inaccurately estimated PSD, together with imperfect centering of particles and the ignored astigmatism in CTF, may cause imperfect CTF correction and additional blurring in the restored images. However, we demonstrate in our experimental results that our methods are more robust to these factors than the original CWF, especially for contrast estimation. {\color{black} The machine and the number of cores we used for the experimental datasets are the same as those of the synthetic simulations.}

\subsection{EMPIAR-10028}\label{sec:10028}
We test the algorithms on {\color{black}a} dataset of the Plasmodium falciparum 80S ribosome bound to the anti-protozoan drug emetine. The picked particles are downloadable from EMPIAR-10028 \cite{10028}. Its
3-D reconstruction can be found on EMDB as EMD-2660 \cite{10028}. {\color{black}The dataset} consists of 105247 motion corrected and picked particle images of size $360\times 360$ with 1.34 \AA~pixel
size, from 1081 defocus groups.
{\color{black}
We estimate the covariance using all images, and use 21 defocus groups to estimate the contrast of individual images and then denoise the selected images.}
The background subtraction, whitening, Fourier-Bessel expansion and covariance estimation  {\color{black} took 10} hours. It  {\color{black} took 5 seconds} for SDP covariance refinement and {\color{black}less than $1$} second for the GS one. We apply Wiener filtering to 21 defocus groups, which take {\color{black}$11$ minutes}. Contrast estimation  from the Fourier-Bessel coefficients took less than one second.

{\color{black}
We first examine the relationship between the contrasts of particle images and their locations in a micrograph. Since the ground truth clean contrasts are not available, we use the approximate ground truth contrasts that are obtained from template matching with clean projections of the 3-D volume estimated by RELION (available in EMD-2660). In order to do this, we first generate 1000 clean templates that are projected from uniformly distributed viewing directions. Next, for each particle image, we find its viewing direction, 2-D in-plane rotation and shift by aligning its CWF-denoised image with each of the clean template by the method of \cite{reddy1996fft}. We found that using the denoised images often provide more accurate alignment than using the raw images. To compute the oracle contrast of each noisy image $\bY_i$, we apply its CTF to the aligned clean template and obtain $\bY_i^*$. However, the contrast directly computed by
$$
    c_i^* = \langle\bY_i\,, \bY_i^*\rangle/\|\bY_i^*\|^2_F
$$
can be sensitive to even slight errors in alignment, as we demonstrated in Figures \ref{fig:contrast_white_df} and \ref{fig:contrast_color_df} in synthetic data simulations. To mitigate this issue,
we apply Gaussian smoothing to both $\bY_i$ and $\bY_i^*$ before computing the ``ground truth" contrast using the above formula. We choose an envelope function with a $B$-factor $1000$ as our Gaussian filter. We remark that the clean projections are only used to generate approximate ground truth for evaluation, and are not used in CWF and our methods.

Each subplot of Figure \ref{fig:relion_coord_10028} corresponds to one micrograph, where each dot represents a picked particle image in that micrograph. The location of the dots are the location of the particle images in that micrograph, whose color represents the oracle contrast by template matching. The defocus values of the three micrographs (from left to right) are respectively 0.8131 {\color{black}\si{\mu m}}, 1.9676 {\color{black}\si{\mu m}}, and 2.6643 {\color{black}\si{\mu m}}. Figure \ref{fig:relion_coord_10028} suggests the existence of local correlations of image contrasts. Indeed, many pairs of nearby particles have very similar contrasts. However, the correlation of contrasts is only present within very small sub-regions of the micrograph.
\begin{figure}[H]
    \centering
    \includegraphics[width=1\columnwidth]{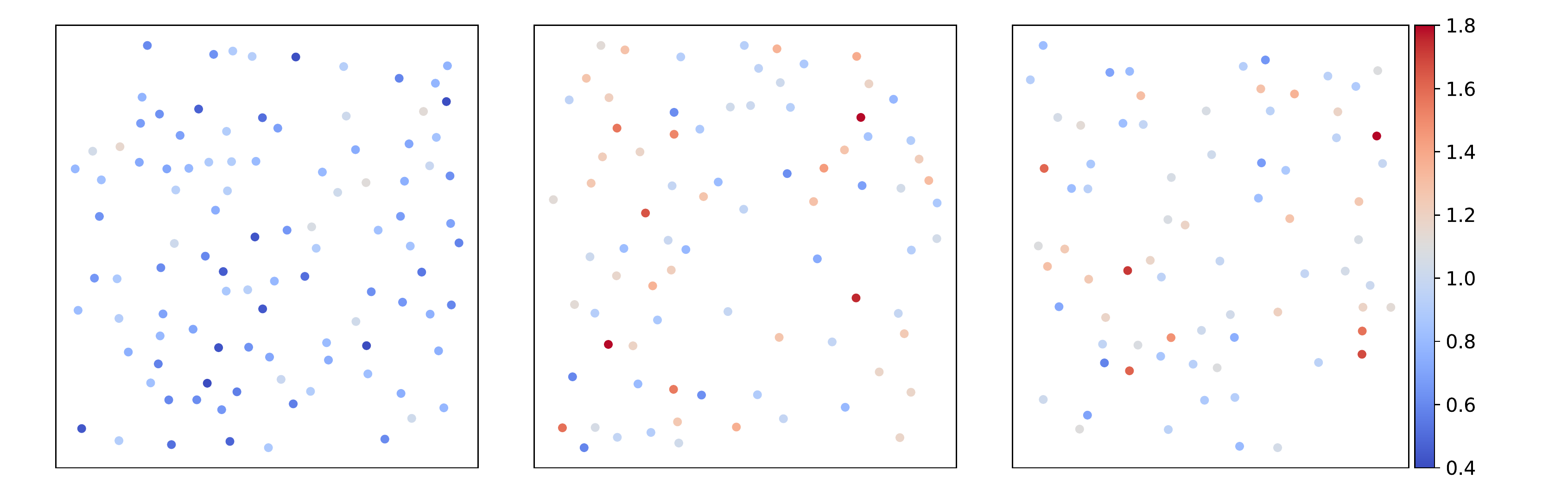}
    \caption{{\color{black}Demonstration of the relationship between the contrast of picked particle  and their locations in the micrographs of EMPIAR-10028. Each dot corresponds to a particle image, whose color represents its estimated contrast.}}
    \label{fig:relion_coord_10028}
\end{figure}

We next present a box plot of both oracle contrasts (top subplot) and the contrasts estimated by CWF-GS (bottom subplot) for each of the 21 defocus groups in Figure \ref{fig:box_10028}. We ignore the result from CWF-SDP as it is very similar to the one of CWF-GS. We also ignore the result from CWF, since its contrast estimation is not accurate (see later in Figure \ref{fig:scatter_10028}) and thus its box plot is not informative. From left to right in each subfigure, the defocus values are sorted in ascending order, ranging from 0.8131 {\color{black}\si{\mu m}} to 2.6643 {\color{black}\si{\mu m}}. In each box plot, the 5 horizontal lines, from top to bottom, respectively correspond to max value, $75\%$ quantile, median,  $25\%$ quantile and min value. The two box plots are similar, even though their contrasts are estimated using completely different methods. One can also see a clearer trend from the second subfigure (our method) that micrographs with higher defocus values tend to have higher contrast. This makes sense, as CTFs with higher defocus values preserve more low frequency information, which yields higher SNR in low frequencies.
Interestingly, both subfigures show that contrast variation within each micrograph is often larger than the variance of the median contrast of each micrograph (the variance of the y-values of the orange lines). This possibly indicates that using a single contrast value per micrograph, as assumed in the 3-D iterative refinement stage, is not appropriate.

Next, we present the scatter plot between the estimated contrasts and the oracle contrast.
\noindent It is clear from Figure \ref{fig:scatter_10028} that our estimates have much better correlation with the oracle. {\color{black}We remark that we do not expect a strong correlation in any case, since the oracle itself is noisy and suffers from imperfect alignment. However, this is strong evidence that our methods provide much better contrast estimates than CWF. }
\begin{figure}[!htbp]
    \centering
    \includegraphics[width=1\columnwidth]{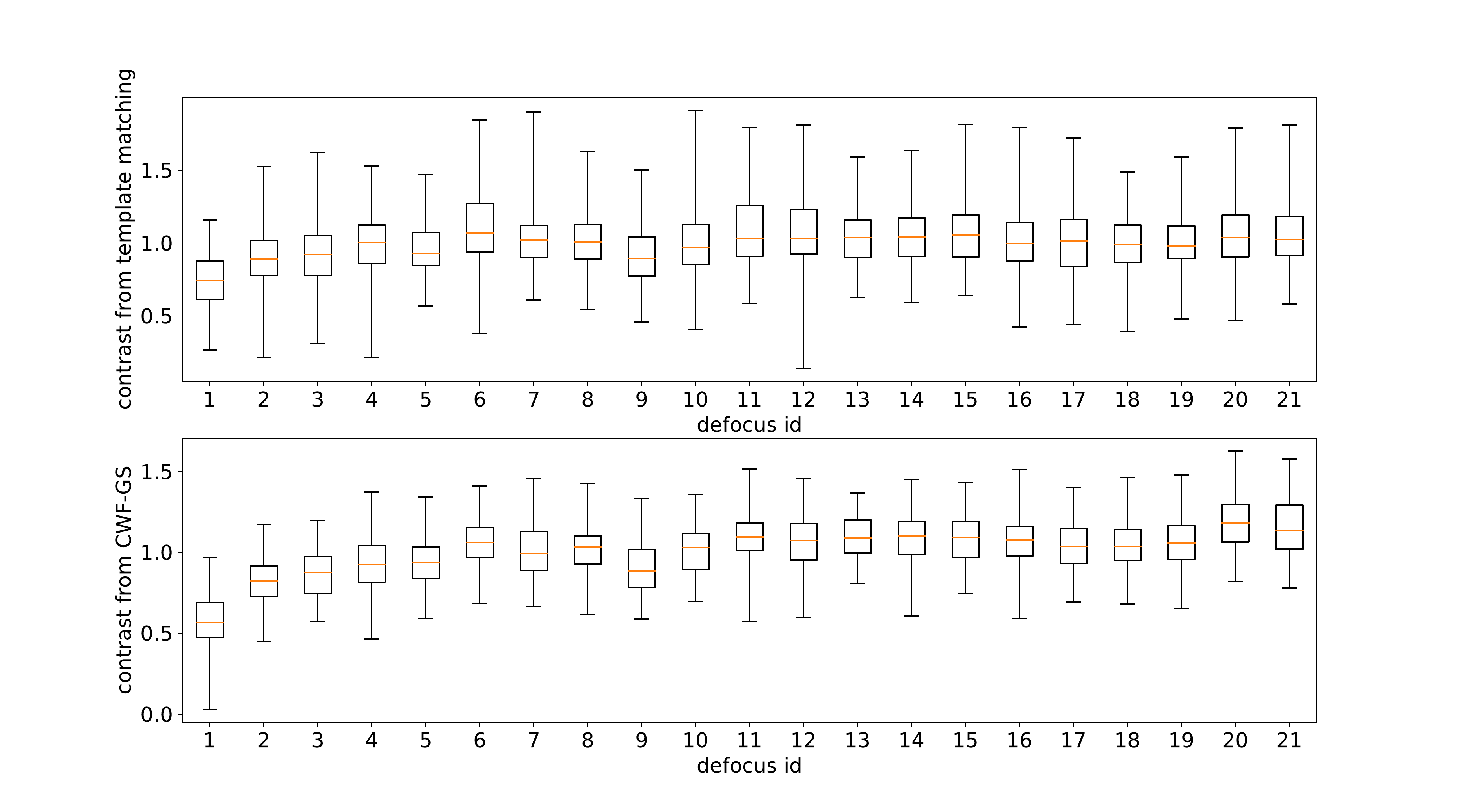}
    \caption{\color{black}Box plot of the oracle contrasts (top) and our estimated contrasts (bottom) in 21 defocus groups of the dataset EMPIAR-10028. The defocus values are sorted in ascending order, ranging from 0.8131 {\color{black}\si{\mu m}} to 2.6643 {\color{black}\si{\mu m}}. }
    \label{fig:box_10028}
\end{figure}

\begin{figure}[!htbp]
    \centering
    \includegraphics[width=1\columnwidth]{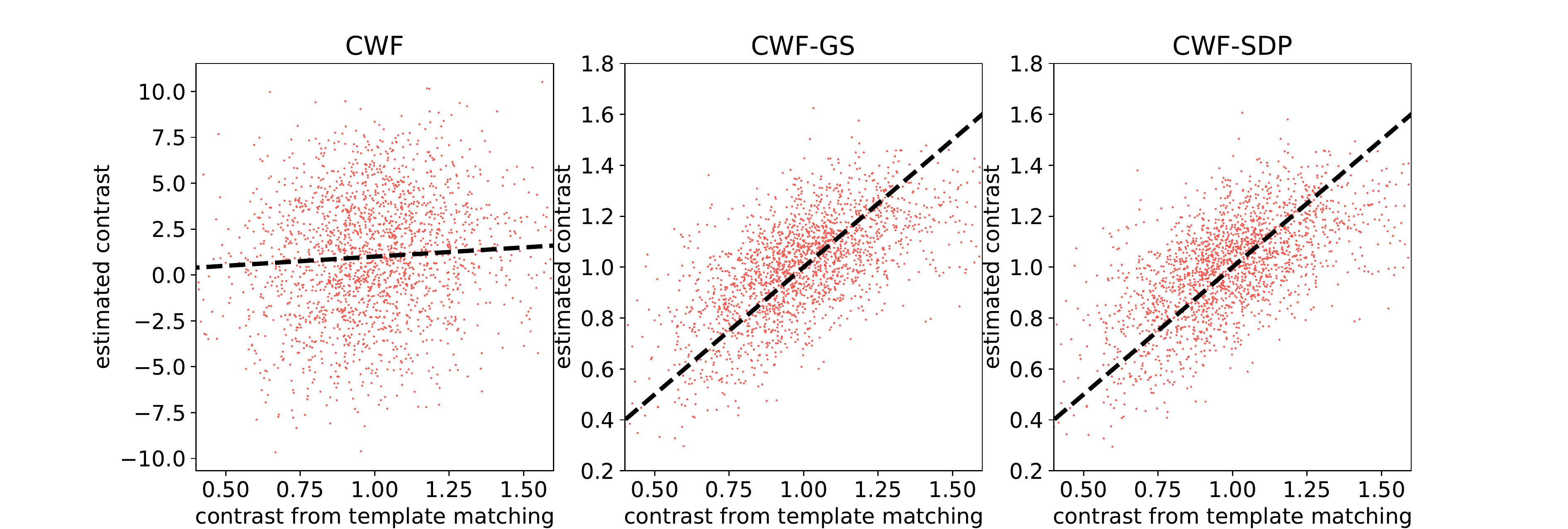}
    \caption{The scatter {\color{black}plots} of the estimated contrast v.s. the oracle contrast for three defocus groups in the dataset EMPIAR-10028. {\color{black} The dashed line corresponds to the function $y=x$.} }
    \label{fig:scatter_10028}
\end{figure}

}

\begin{figure}[H]
    \centering
    \includegraphics[width=0.8\columnwidth]{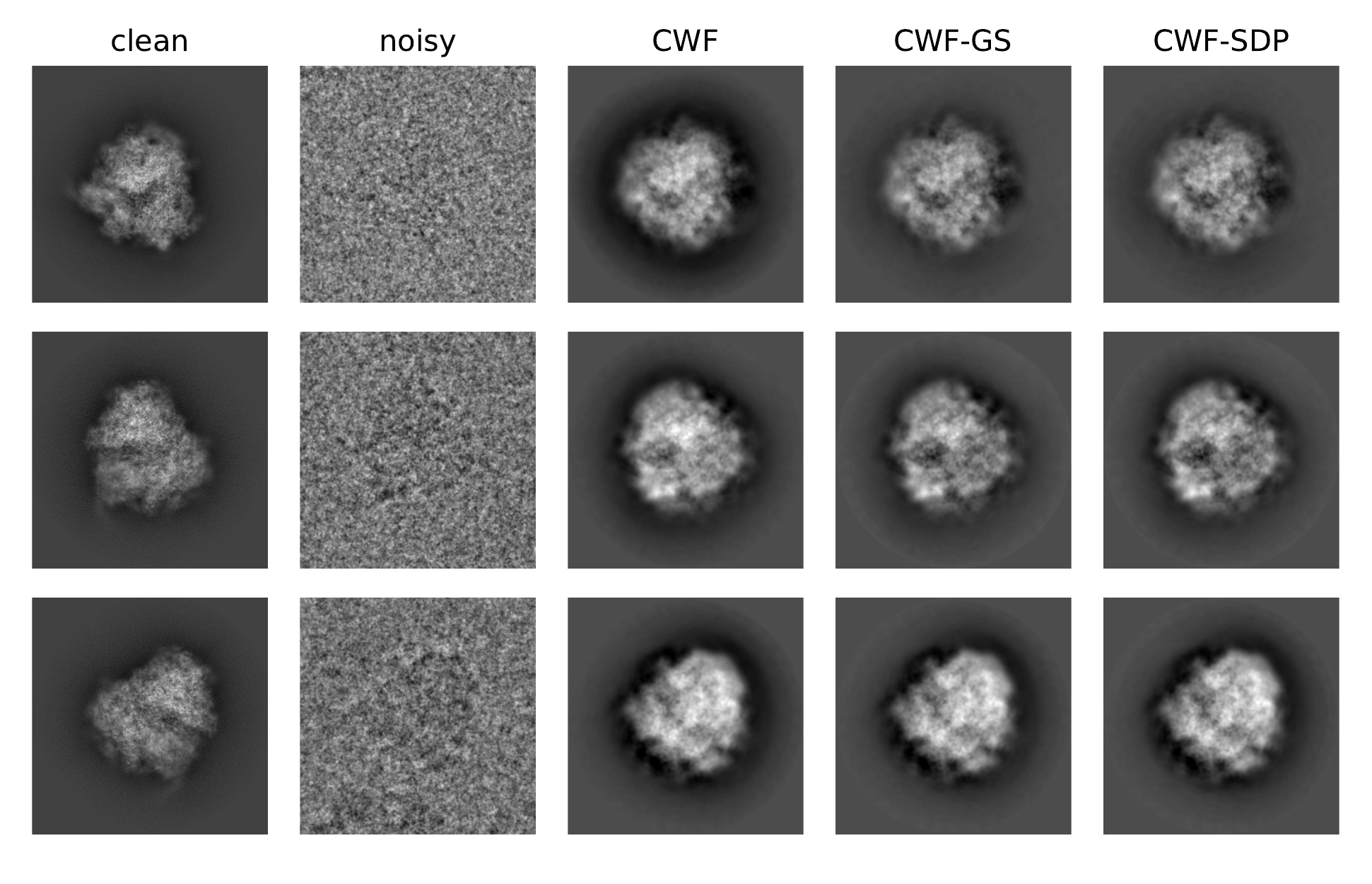}
    \caption{{\color{black} Denoising} results of EMPIAR-10028.}
    \label{fig:denoise_10028}
\end{figure}

We next compare the image denoising performance by CWF and the ones with our refined covariance matrix. Since image normalization only affects the global scale of the image and normalized CWF performs poorly, we only show the denoised images without normalization. We also found that the 2-stage CWF often performs worse than the 1-stage version, possibly due to violation of assumptions in our synthetic model, such as imperfect centering and astigmatism in CTF. Thus, we recommend applying the one-stage algorithm for experimental datasets, and we compare their denoised images as follows.
From Figure \ref{fig:denoise_10028}, all methods produce dark areas around the boundary of the particle. These dark rings are likely due to the imperfect CTF correction by CWF. The denoised images by our methods have less dark areas, comparing to that of CWF. {\color{black} Since negative pixel values are often observed in CTF-affected clean images, these dark rings in CWF-denoised images are possibly due to inaccurate CTF-correction, which suggests better CTF correction by our methods. Furthermore, we observe better denoised images by our methods with closer contrast to the clean templates.}

{\color{black}At last, to quantitatively compare the denoising results of different methods, we compute the Fourier ring correlations (FRC) between the denoised images and their aligned clean templates. That is, for each pair of two images $I_1$ and $I_2$ and their Fourier coefficient vectors $\boldsymbol f_{1,r}, \boldsymbol f_{2,r}$ at radial frequency $r$, we compute
$$
\text{FRC}(r) = \frac{\Re(\boldsymbol f_{1,r}^* \boldsymbol f_{2,r})}{\|\boldsymbol f_{1,r}\|\|\boldsymbol f_{2,r}\|},
$$
where $\Re$ denotes taking the real part of a complex number.
For each method, we compute the average FRC between the denoised images and the clean templates over the 2015 images from 21 defocus groups.
We notice that FRC is very sensitive to image rotations and shifts. With slight error in image alignment, the FRC of all methods decreases rapidly as $r$ increases. As a result, when alignment errors are present, FRC may not reflect the true image quality. However, even from the first few frequencies, the FRCs of CWF-based methods are much higher than that of the na\"ive phase flipping method. We also notice that CWF-denoised images have large errors in the first two frequencies, mainly due to its limitations in handling contrast variations. Since the clean templates are only aligned and registered with CWF-denoised images, the comparison is a bit unfair to our methods, as our methods may suffer from larger alignment errors. However, even in this scenario, our methods achieve much better FRC at the first two radial frequencies due to the better contrast estimation. 

}
\begin{figure}[H]
    \centering
    \includegraphics[width=0.7\columnwidth]{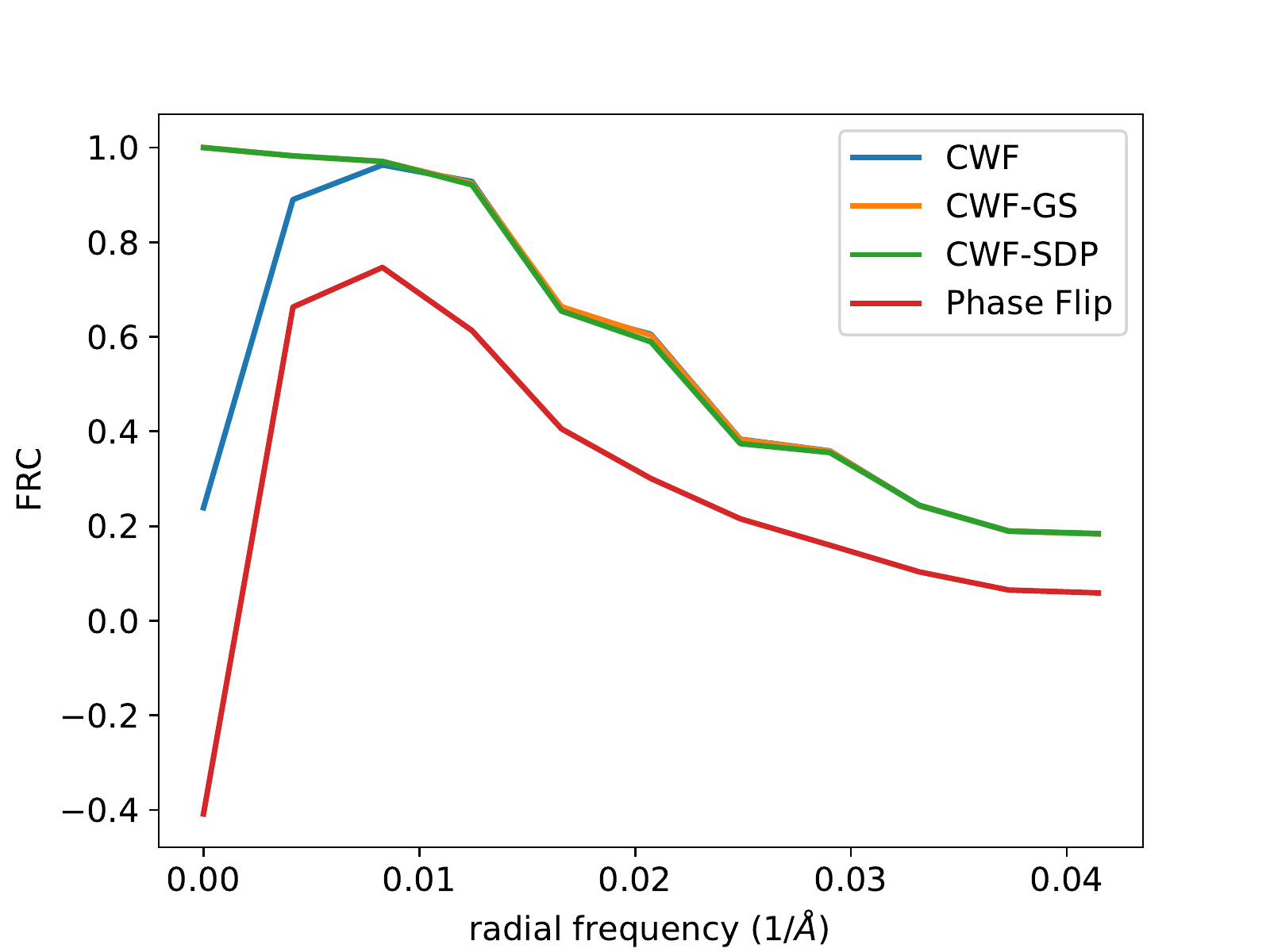}
    \caption{\color{black}The average Fourier ring correlation between denoised images and the aligned clean templates over 2015 images from EMPIAR-10028.}
    \label{fig:frc_10028}
\end{figure}

\section{Conclusion}
 We introduced an effective algorithm for {\color{black}estimating the amplitude contrast of individual images and the overall contrast variability in the ab-initio stage. Our method refines the initial estimated covariance so it satisfies} additional constraints that follow from the image formation model by tomographic projection. Results for both synthetic and experimental datasets indicate consistently better contrast estimation by our methods than CWF. On synthetic data, the contrast estimation errors of our methods are comparable to those of an oracle, even with small number of images. We {\color{black} also} demonstrate that our method improves the image denoising result of CWF. {\color{black}Among the various contrast estimation and image denoising techniques that were considered in this paper, following the results for experimental datasets we recommend using CWF-GS (see Algorithm \ref{alg:ACE} with option=GS in Section \ref{sec:summary}) for contrast estimation and CWF-GS with image normalization for image denoising (see Algorithm \ref{alg:AD} with option=normalization in Section \ref{sec:summary}).} There are also some interesting future directions. For example, one can try techniques based on common-lines to directly estimate the rotations of molecules by using the denoised and normalized images from our methods with rudimentary 2-D class averaging \cite{bandeira2020non}. Normalizing the images may also lead to improvement of 2-D class averaging procedures.  Another interesting application of our method is to use our estimated contrasts to initialize their values in the iterative refinement procedure of RELION \cite{scheres2012relion}. {\color{black}As for the computational aspect, one can modify the original CWF method so it can more efficiently handle per-image CTF, rather than a small number of defocus groups.}  Our {\color{black}Python} code is available at \url{https://github.com/yunpeng-shi/contrast-cryo} which is planned to be integrated into ASPIRE \cite{aspire}.

\section*{Acknowledgement}
A.S. and Y.S. are supported in part by AFOSR FA9550-20-1-0266, the Simons Foundation Math+X Investigator Award, NSF BIGDATA Award IIS-1837992, NSF DMS-2009753, and NIH/NIGMS
1R01GM136780-01. {\color{black}
We thank the referees and the editor for their valuable comments.} We thank Garrett Wright for his continuous efforts on improving and optimizing the ASPIRE package, especially for his work on correcting and improving the code of CWF. We also thank Chris Langfield for his generous help on cleaning and fixing the star files of the experimental datasets. At last, we thank Eric J. Verbeke for valuable discussions. 

\appendix

\section{Additional Results for Experimental Data}
This appendix provides the results of applying our methods for contrast estimation and image denoising for two experimental datasets. Overall, the results for these two datasets are similar to the results with the experimental dataset reported in the main body text.

\subsection{Results for EMPIAR-10005}
We compare our algorithms using an experimental dataset of the TRPV1 ion channel, whose ID is EMPIAR-10005 \cite{10005}. Its estimated
3-D volume is also available on EMDB with ID EMD-5778 \cite{10005}. The dataset contains 35645 picked particle images of
size $256\times 256$ with pixel size of 1.2156 \AA~ from 935 defocus groups.

We estimate the covariance using all images, and pick {\color{black}19 defocus groups (0th, 50th, 100th, 150th, ..., 900th)} to estimate the contrast of individual images and then denoise the selected images. The background subtraction, whitening, Fourier-Bessel expansion and covariance estimation takes {\color{black}4} hours. It takes {\color{black}1.5 seconds} for SDP covariance refinement and {\color{black}less than 1 second} for the GS one. Applying Wiener filtering to the {\color{black}19} defocus groups takes {\color{black} 5 minutes}. The time for contrast estimation  from the Fourier-Bessel coefficients is less than one second.

\begin{figure}[H]
    \centering
    \includegraphics[width=0.8\columnwidth]{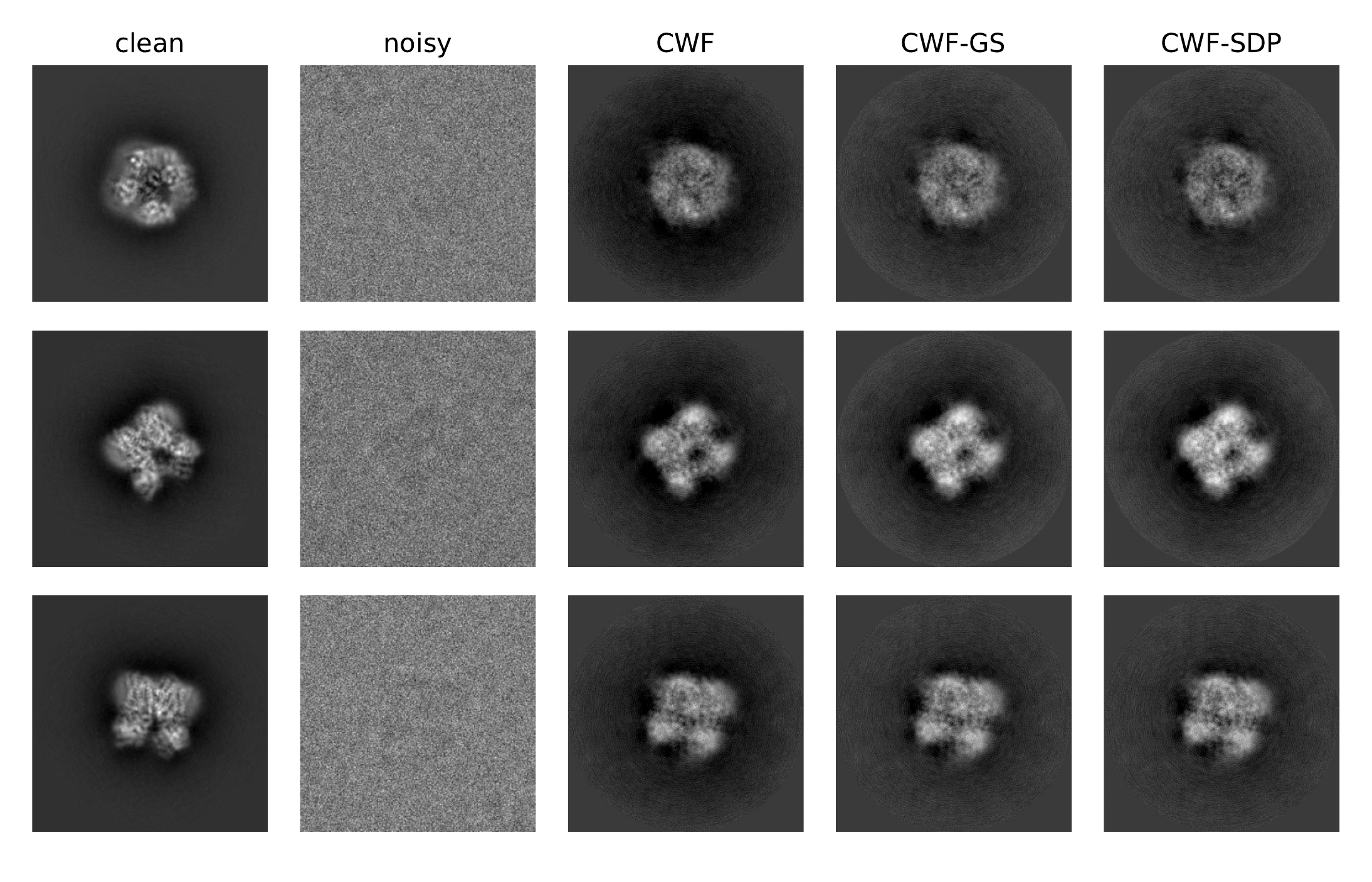}
    \caption{{\color{black} Denoising} results of EMPIAR-10005.}
    \label{fig:denoise_10005}
\end{figure}

{\color{black}
We first present a box plot of both oracle contrasts (top subplot) and the contrasts estimated by CWF-GS (bottom subplot) for each of the 19 defocus groups in Figure \ref{fig:box_10005}. From left to  right in each subfigure, the defocus values are sorted in ascending order, ranging from 0.9889 $\mu m$ to 2.1976 $\mu m$. In each box plot, the 5 horizontal lines, from top to bottom, respectively correspond to max value, $75\%$ quantile, median,  $25\%$ quantile, and min value. Similar to EMPIAR-10028, the box plots from the two subfigures are similar, and both subfigures show large contrast variation within each micrograph. However, from the box plots we do not see a strong correlation between the defocus values and the image contrasts, unlike the results for EMPIAR-10028.

Next, we present the scatter plot between the estimated contrasts and the oracle contrast.
\noindent It is clear from Figure \ref{fig:scatter_10005} that our estimates have much better correlation with the oracle. In particular, CWF-GS performs slightly better than CWF-SDP. The scatter plots of both methods tend to have slope smaller than 1, due to the low SNR of this dataset.
\begin{figure}[!htbp]
    \centering
    \includegraphics[width=1\columnwidth]{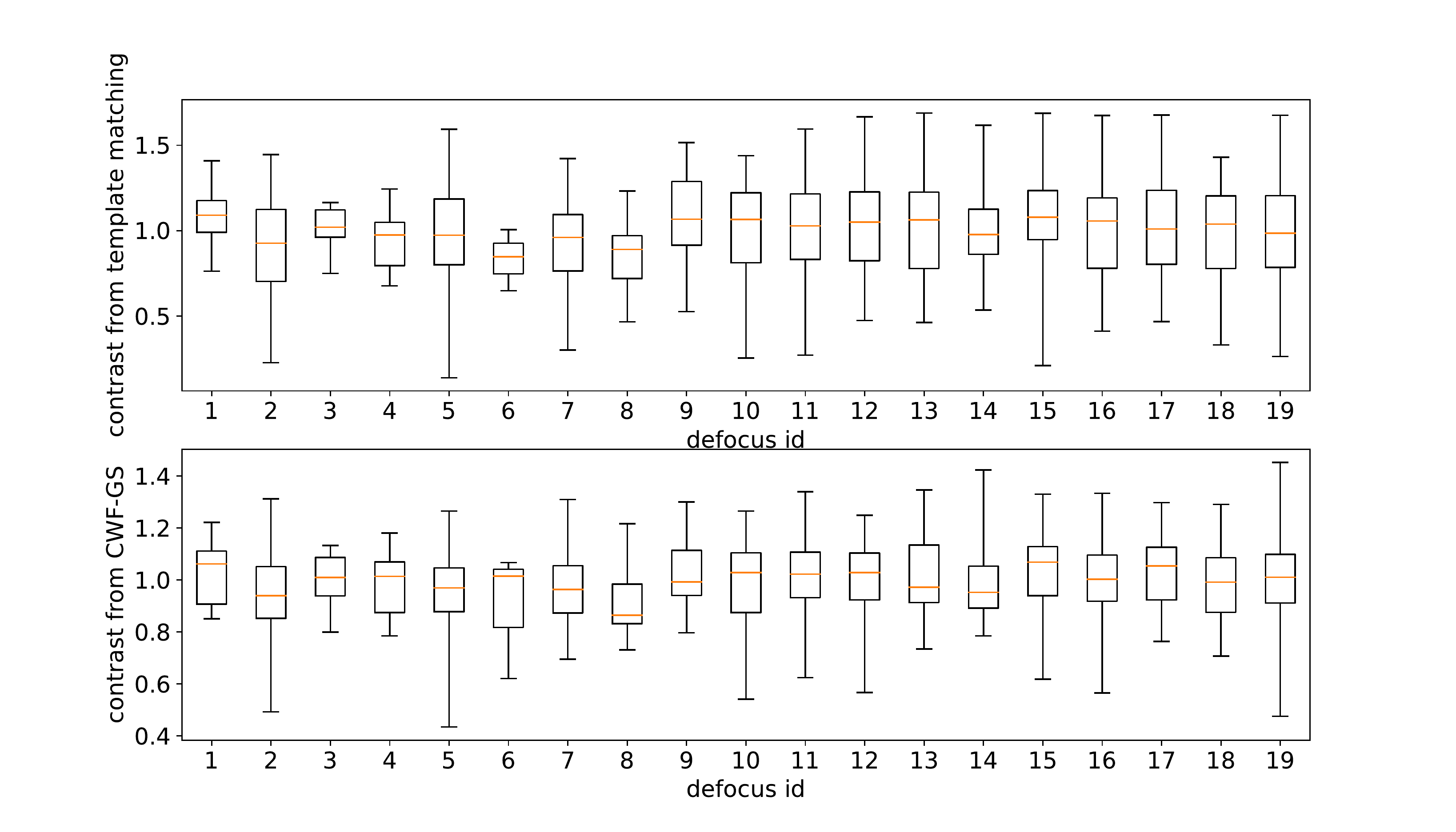}
    \caption{\color{black}Box plot of the oracle contrasts (top) and our estimated contrasts (bottom) in 19 defocus groups of the dataset EMPIAR-10005. The defocus values are sorted in ascending order, ranging from 0.9889 $\mu m$ to 2.1976 $\mu m$. }
    \label{fig:box_10005}
\end{figure}

\begin{figure}[!htbp]
    \centering
    \includegraphics[width=1\columnwidth]{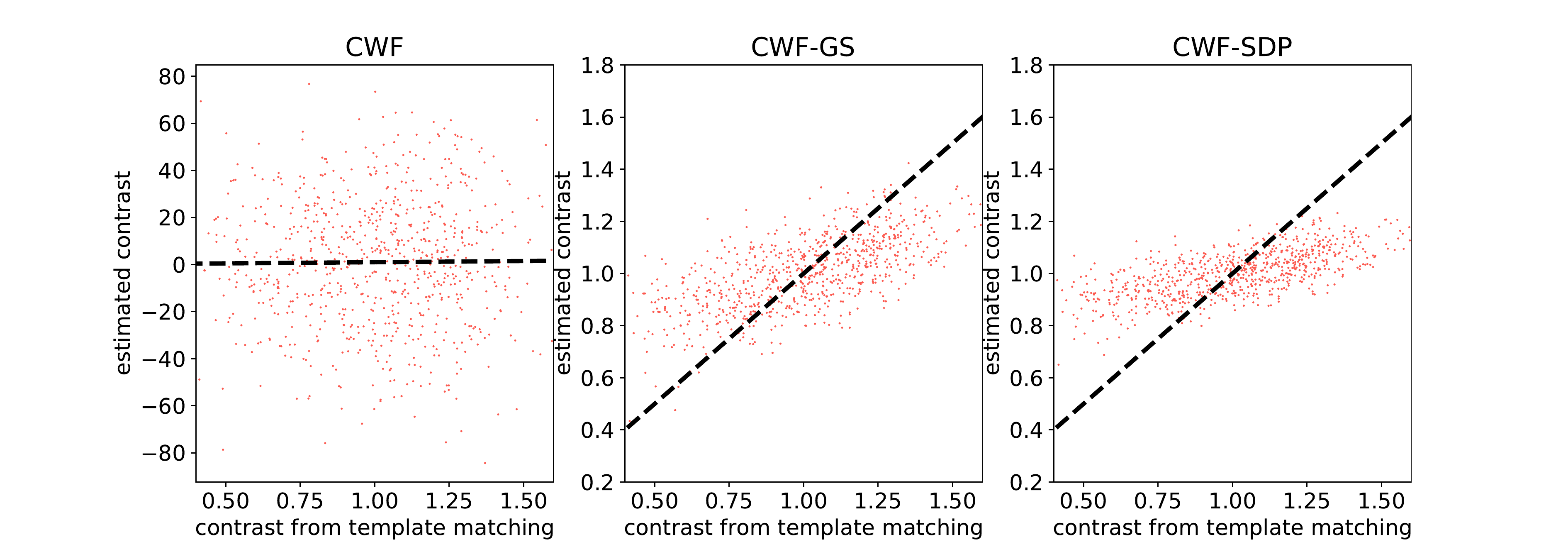}
    \caption{Scatter plot of the estimated contrast v.s. the oracle contrast for three defocus groups in the dataset EMPIAR-10005. {\color{black}The dashed lines correspond to the function $y=x$.}}
    \label{fig:scatter_10005}
\end{figure}

}
We next compare the image denoising performance by CWF and the ones with our refined
covariance matrix. From Figure \ref{fig:denoise_10005}, we {\color{black}observe} similar dark rings as in the example of EMPIAR-10028
for denoised images by CWF. This issue is largely mitigated by applying our covariance
refinement methods. 

\begin{figure}[H]
    \centering
    \includegraphics[width=0.7\columnwidth]{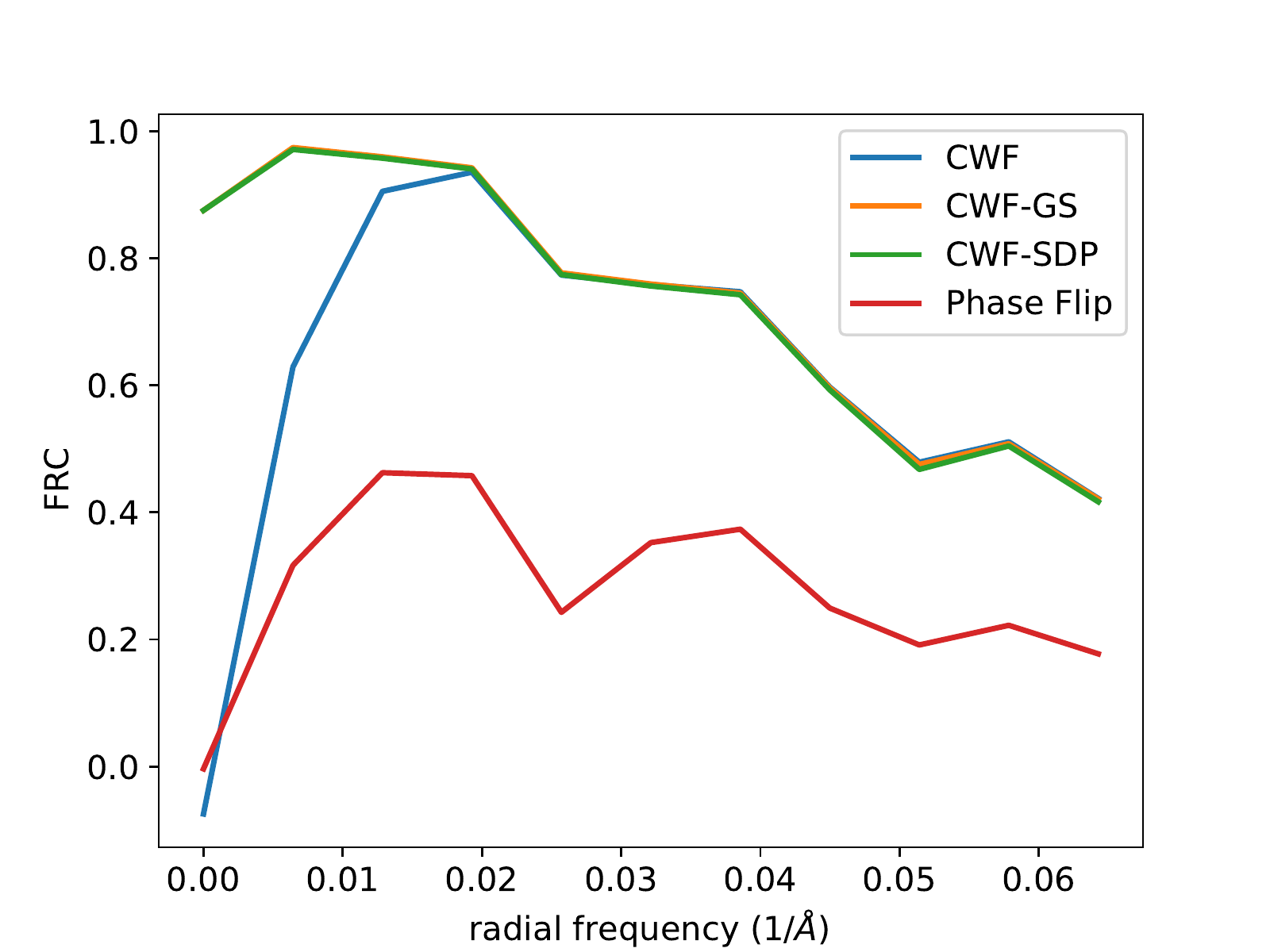}
    \caption{\color{black}The average FRC between denoised images and the aligned clean templates over 817 images from EMPIAR-10005.}
    \label{fig:frc_10005}
\end{figure}

{\color{black}At last, to quantitatively compare the denoising results of different methods, we compute the FRCs between the denoised images and their aligned clean templates (Figure \ref{fig:frc_10005}). 
For each method, we compute the average FRC between the denoised images and the clean templates over  817 images from 19 defocus groups.
Again, the FRCs of CWF-based methods are much higher than that of the na\"ive  phase flipping method, and our methods achieve much better FRC than other methods at the first few radial frequencies due to the superior contrast estimation. 

}

\subsection{Results for EMPIAR-10073}
We test the algorithms on the dataset of the yeast U4/U6.U5 tri-snRNP. The picked particles are downloadable from EMPIAR with ID EMPIAR-10073 \cite{10073}. Its
3-D reconstruction can be found on EMDB as EMD-8012 \cite{10073}. It consists of 138899 motion corrected and picked particle images of size $380\times 380$ with 1.45 \AA~pixel
size, from 2340 defocus groups.

We estimate the covariance using all images, and pick {\color{black}21 defocus groups (0th, 100th, 200th, ..., 2000th)} to estimate the contrast of individual images and then denoise the selected images.
The background subtraction, whitening, Fourier-Bessel expansion and covariance estimation takes {\color{black}20} hours. In the current implementation of CWF the running time is proportional to the number of defocus groups, which is the reason why this step takes longer for this dataset. It takes 
{\color{black}7 seconds} for SDP covariance refinement and {\color{black}less than 1 second} for the GS one. Applying Wiener filtering to the {\color{black}$21$} defocus groups takes {\color{black}$9$ minutes}. The time for contrast estimation  from the Fourier-Bessel coefficients is less than one second.

{\color{black}
We first present a box plot of both oracle contrasts (top subplot) and  contrasts estimated by CWF-GS (bottom subplot) for each of the 21 defocus groups in Figure \ref{fig:box_10073}. From left to  right in each subfigure, the defocus values are sorted in ascending order, ranging from 0.4609 $\mu m$ to 2.4008 $\mu m$. Similar to the previous example, we do not observe a clear correlation between defocus values and image contrast, and
the contrast variation within each micrograph is large.

Next, we present the scatter plot of the estimated contrasts v.s. the oracle contrast.
\noindent From Figure \ref{fig:scatter_10073}, our estimates have better correlation with the oracle. {\color{black}However, due to the low SNR of EMPIAR-10073, our methods perform worse than in previous examples.}
\begin{figure}[!htbp]
    \centering
    \includegraphics[width=1\columnwidth]{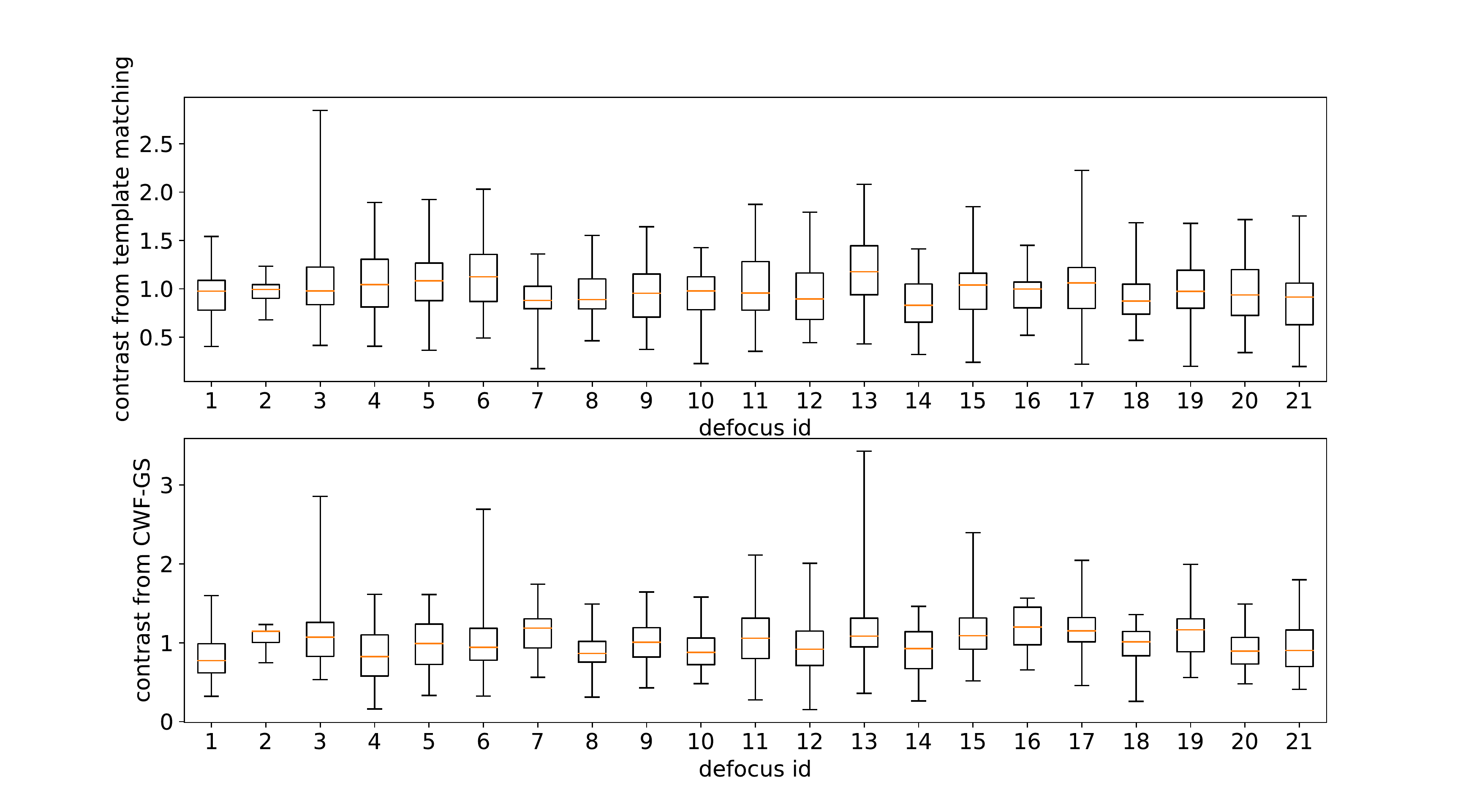}
    \caption{\color{black}Box plot of the oracle contrasts (top) and our estimated contrasts (bottom) in 21 defocus groups of the dataset EMPIAR-10073. The defocus values are sorted in ascending order, ranging from 0.4609 $\mu m$ to 2.4008 $\mu m$. }
    \label{fig:box_10073}
\end{figure}

\begin{figure}[!htbp]
    \centering
    \includegraphics[width=1\columnwidth]{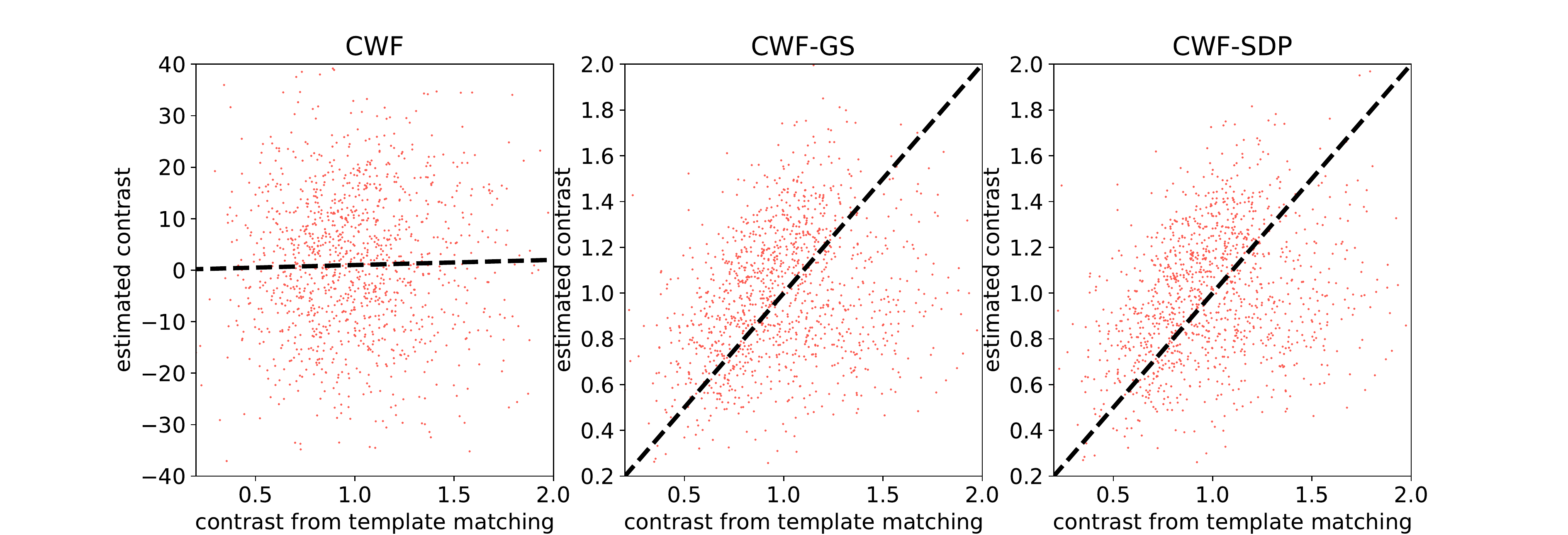}
    \caption{Scatter plot of the estimated contrast v.s. the oracle contrast for three defocus groups in the dataset EMPIAR-10073. {\color{black}The dashed lines correspond to the function $y=x$.}}
    \label{fig:scatter_10073}
\end{figure}

}

\begin{figure}[H]
    \centering
    \includegraphics[width=0.8\columnwidth]{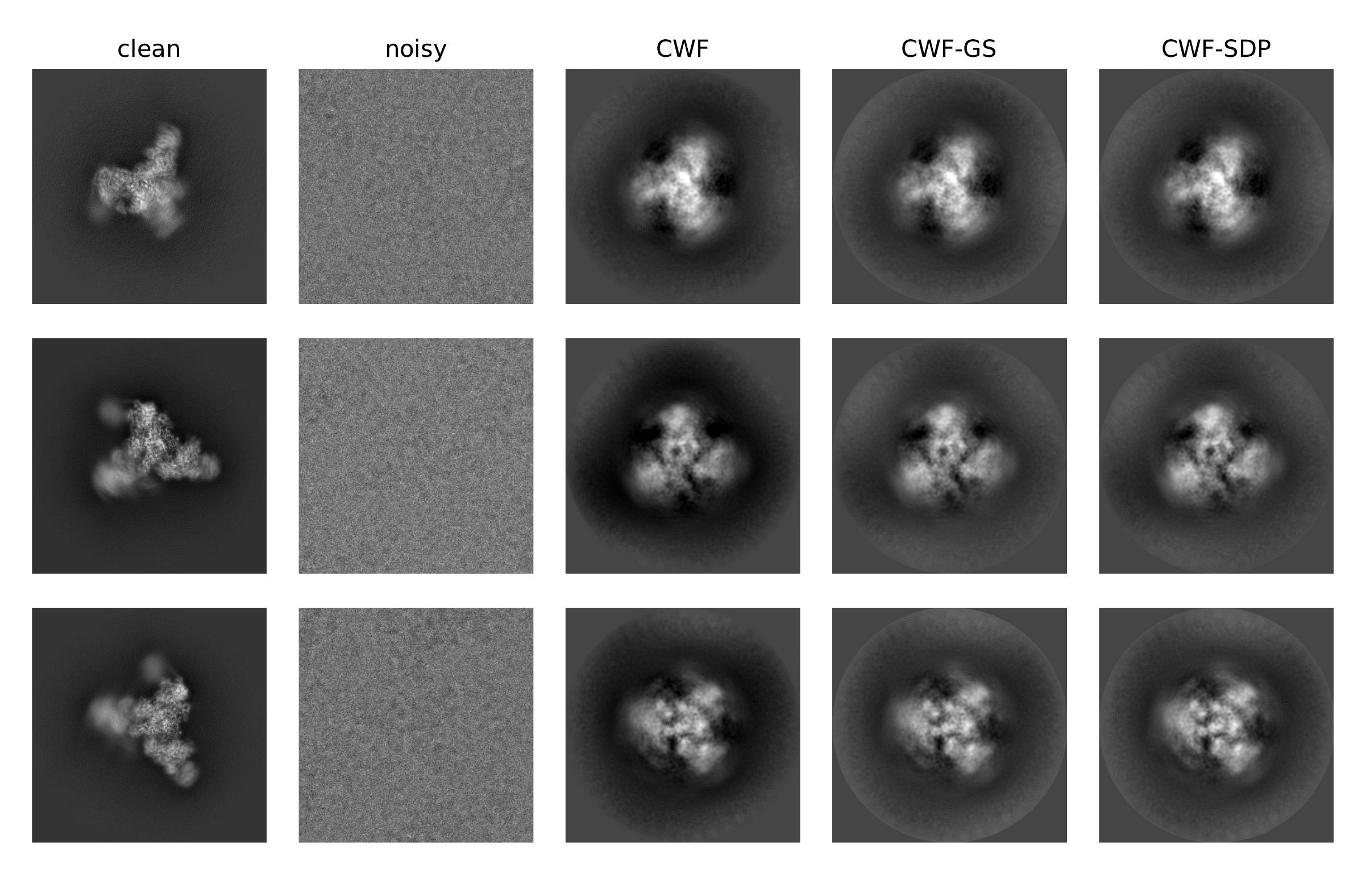}
    \caption{{\color{black} Denoising} results of EMPIAR-10073.}
    \label{fig:denoise_10073}
\end{figure}

We next compare the image denoising performance by CWF with our refined covariance matrix. 
From Figure \ref{fig:denoise_10073}, the denoised images by our methods have less dark areas, comparing to that of CWF, which suggests better CTF correction by our methods.

{\color{black}At last, we present the FRCs between the denoised images and their aligned clean templates. 
For each method, we compute the average FRC between the denoised images and the clean templates over  1206 images from 21 defocus groups. Similar to the previous datasets, the denoised images by our methods have higher FRCs at the first few frequencies, compared to those of CWF.
}
\begin{figure}[H]
    \centering
    \includegraphics[width=0.7\columnwidth]{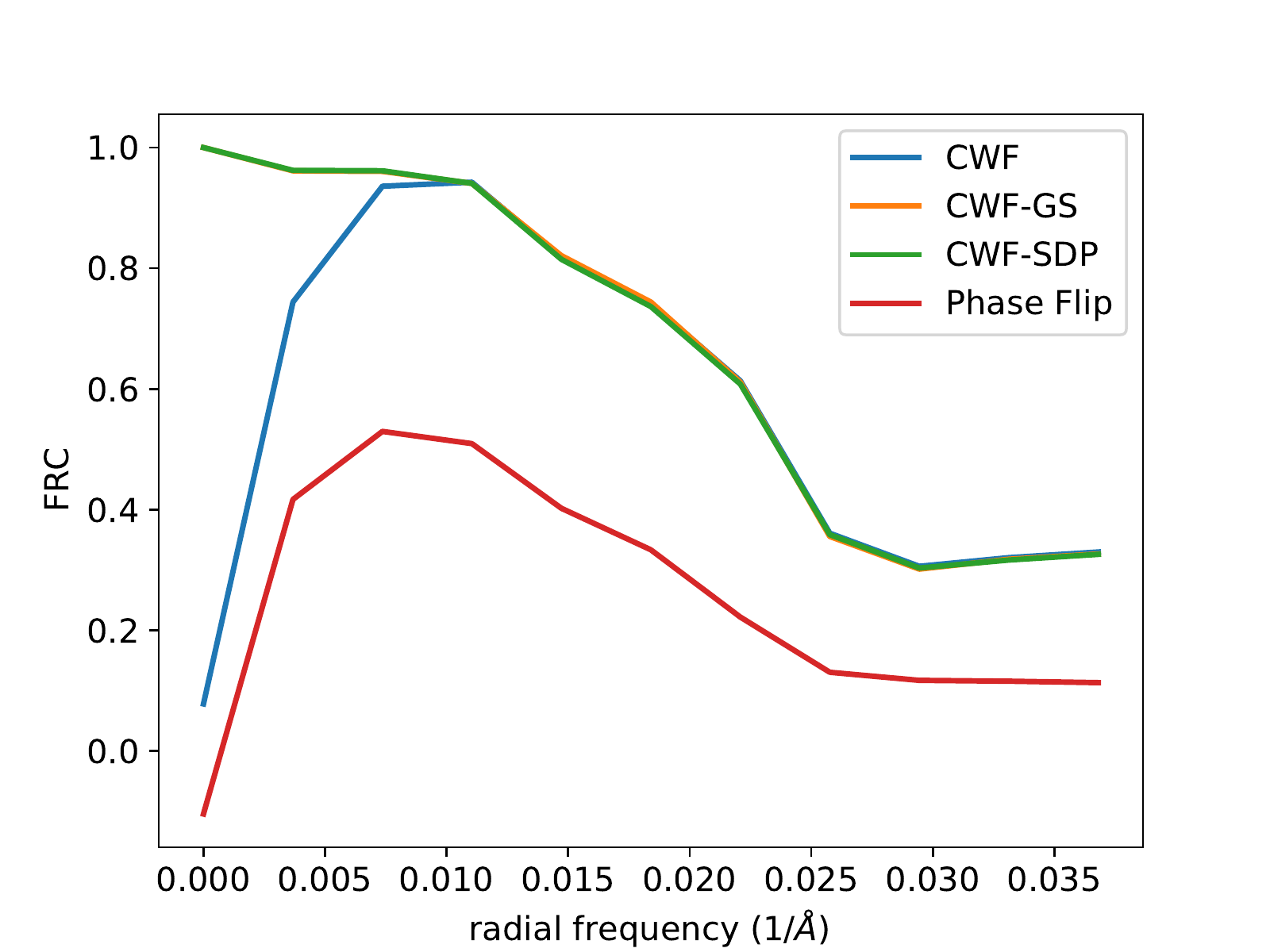}
    \caption{\color{black}The average FRC between denoised images and the aligned clean templates over 1206 images from EMPIAR-10073.}
    \label{fig:frc_10073}
\end{figure}

{\small
\bibliographystyle{abbrv}
\bibliography{cryo_bib}
}

\end{document}